\documentclass[letterpaper, 10 pt, conference]{ieeeconf}
\IEEEoverridecommandlockouts
\usepackage{amsmath,amsfonts}
\usepackage{amsmath} % for 'bmatrix' environment
\usepackage{newtxtext,newtxmath} % optional -- Times Roman clone
\usepackage{algorithmic}
\usepackage{algorithm}
\usepackage{array}
\usepackage[caption=false,font=footnotesize,labelfont=sf,textfont=sf]{subfig}
\usepackage{textcomp}
\usepackage{stfloats}
\usepackage{url}
\usepackage{verbatim}
\usepackage{graphicx}
\usepackage{cite}
\usepackage{hyperref}
\usepackage[table,xcdraw]{xcolor}
\usepackage{bm}
\usepackage{bbm}
\usepackage{subfig}
\usepackage{textcomp}
\usepackage{accents}
\usepackage{multirow}
\title{\LARGE \bf A Generic Trajectory Planning Method for \\ Constrained All-Wheel-Steering Robots}
\author{
Ren Xin, Hongji Liu, Yingbing Chen, Jie Cheng, Sheng Wang, Jun Ma, and Ming Liu
\thanks{This work was supported in part by the Guangzhou-HKUST(GZ) Joint Funding Scheme under Grants 2023A03J0148 and 2024A03J0618. \textit{(Corresponding Author: Jun Ma.)}}
\thanks{Ren Xin, Hongji Liu, Yingbing Chen, and Sheng Wang are with the Division of Emerging Interdisciplinary Areas, The Hong Kong University of Science and Technology, Hong Kong SAR, China (email: \{rxin,hliucq,ychengz,swangei\}@connect.ust.hk).}
\thanks{Jie Cheng is with the Department of Electronic and Computer Engineering, The Hong Kong University of Science and Technology, Hong Kong SAR, China (email: jchengai@connect.ust.hk).}
\thanks{Jun Ma and Ming Liu are with the Robotics and Autonomous Systems Thrust, The Hong Kong University of Science and Technology (Guangzhou), Guangzhou, China  (email: jun.ma@ust.hk; eelium@hkust-gz.edu.cn).
}
}
\begin{document}
\maketitle

\begin{abstract}
This paper presents a generic trajectory planning method for wheeled robots with fixed steering axes while the steering angle of each wheel is constrained. 
In the existing literatures, All-Wheel-Steering~(AWS) robots, incorporating modes such as rotation-free translation maneuvers, in-situ rotational maneuvers, and proportional steering, exhibit inefficient performance due to time-consuming mode switches. 
This inefficiency arises from wheel rotation constraints and inter-wheel cooperation requirements. 
The direct application of a holonomic moving strategy can lead to significant slip angles or even structural failure. 
Additionally, the limited steering range of AWS wheeled robots exacerbates non-linearity characteristics, thereby complicating control processes.
To address these challenges, we developed a novel planning method termed Constrained AWS~(C-AWS), which integrates second-order discrete search with predictive control techniques. 
Experimental results demonstrate that our method adeptly generates feasible and smooth trajectories for C-AWS while adhering to steering angle constraints.
%  repo link
% git@github.com:Rex-sys-hk/AWSPlanning.git
Code and video can be found at \href{https://github.com/Rex-sys-hk/AWSPlanning}{https://github.com/Rex-sys-hk/AWSPlanning}.

\end{abstract}

% \begin{IEEEkeywords}All-Wheel-Steering, Trajectory planning, Intelligent Vehicle.\end{IEEEkeywords}

\section{Introduction}

\subsection{Motivation}
The benefits of the omnidirectional wheel chassis, particularly during low-speed maneuvers, are well recognized, enabling operations such as spot turning and translation that are unfeasible with traditional platforms with Ackerman geometry~\cite{1642217,9982030}. 
However, complexity and lack of intuitiveness in human-machine interaction for All-Wheel-Steering~(AWS) vehicles have hampered their widespread use~\cite{5153343}.
% due to the challenges they present to operators 

\begin{figure}[t]
    \centering
    \includegraphics[width=\linewidth]{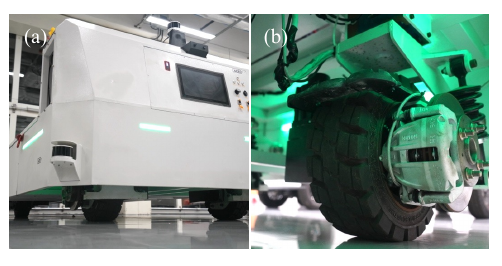}
    \vspace{-2.8em} \\
    \hspace{-0.45em}
    \includegraphics[width=0.955\linewidth]{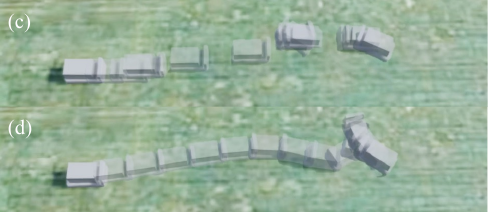}
    \caption{
    % This picture shows our all-wheel steering experimental platform for airport baggage transfer.
(a) A typical autonomous airport baggage carrier, equipped with LiDAR, cameras, computing units, and AWS chassis.
(b) Design of the vehicle's steering wheel. The image reveals that despite the wheel's ability to rotate via the top-mounted steering motor, its free rotation is constrained by essential circuits and hydraulic systems.
(c) and (d) The operation of the AWS vehicle through mode switching and the movement optimization achieved by the proposed C-AWS algorithm, respectively.
    }
    \label{fig:wheel_structure}
    \vspace{-1.5em}
\end{figure}

% The emergence of autonomous driving technologies is enhancing the utility of all-wheel-steering chassis by facilitating their control. 
% Autonomous operation necessitates considering the rotation angles and speeds of all wheels, a complexity not found in the simplified motion control offered by Ackerman geometry~\cite{4059182}. 

The advent of autonomous driving technologies is improving the functionality of all-wheel-steering robots, making their control more manageable. 
Autonomous operation requires consideration of the rotation angles and speeds of all wheels, introducing complexities absent in the simplified motion control of Ackerman geometry~\cite{4059182}.
The concept of the Instantaneous Center of Motion~(ICM)~\cite{pentzer2014model}, is widely employed in describing the motion of rigid bodies, to make sure coherence in the motion states of each wheel.
ICM theory posits that the motion of a rigid body can be simplified as rotational motion around a specific center, characterized by angular velocity and radius. 
% This principle is frequently used in vehicle dynamics. 
% It helps deduce the wheel speed and angle relative to the ground frame.
However, the ICM model exhibits two singular conditions when rotating radius approaching zero or infinity, potentially leading to scenarios where the controller lacks a viable solution or may even crash.
Previous works~\cite{FWIDSMRs, 2Track4WS} avoid singularities by spliting the action space into modes which would prevent the robot from making full use of its flexibility.
Christian \textit{et al.}~\cite{Spherical_ICM} remapped the rotation radius to a periodic function using spherical coordinates to address the \textbf{singularity} problem of the ICM method when the rotating radius approaches infinity. 
Nevertheless, Christian \textit{et al.} only solves the mapping relationship between moving speed, wheel walking speed, and steering angle, but cannot solve the robot motion control problem. 
This method does not ensure that the vehicle can effectively follow a path.
% Although this method solves the problem of singularities in the $r$~-value range and the continuity of the state space, the state space is still non-differentiable. 
% Additionally, remapping $r$ introduces nonlinearity, which can adversely affect the trajectory optimization process. 
% During the process of linearizing the model, we found that the mapping between the first-order state variable of the vehicle and the steering angle and wheel speed does not require the involvement of r, greatly improving the linearity of our model and the success rate of problem-solving.
% Reza++ to be removed
% Reza \textit{et al.}\cite{Reza, Reza++} well defined the path-following problem of AWS with wheel velocity bound. 
% They estimate the maximum speed of a certain wheel to ensure time optimality when following a path and take into account the length of the remaining trajectory to determine whether reducing the distance or the angle error should be prioritized, which is similar to Model Predictive Control~(MPC) but in a closed form. 
% However they failed to deal with acceleration bound, and when tracking certain trajectories that require direction, the wheels will shake frequently with significant acceleration.
Reza \textit{et al.}~\cite{Reza, Reza++} comprehensively defined the path-following problem for AWS robots, considering wheel velocity constraints. 
They estimated the maximum speed for each wheel to achieve time optimality and considered the remaining trajectory length to prioritize either distance minimization or angle error reduction. 
% This approach bears resemblance to Model Predictive Control~(MPC) but is executed in closed form. 
However, their model does not address \textbf{acceleration constraints}, leading to frequent wheel oscillations and significant acceleration when tracking specific directional trajectories.
In addition, AWS platforms have problems with \textbf{steering constraints}.
For example, 
airport baggage carriers are confined to a steering range of $\pm90$\textdegree~\cite{campion1996structural}, attributed to factors like structural durability, hydraulic and electric system requirements, 
as illustrated in Fig.~\ref{fig:wheel_structure}(a),(b).
Traditionally, to circumvent constraints, practitioners treated motion within a specific constraint range as a distinct driving mode and manually switched between these modes, leading to a reduction in movement flexibility for the AWS chassis as demonstrated in Fig.~\ref{fig:wheel_structure}(c).

\subsection{Contribution}

To enhance the operating efficiency of Constrained-AWS~(C-AWS) platforms, smoothness, and reliability under constrained conditions, we adopt a predictive control strategy. 
Our approach involves first identifying a viable trajectory to the target pose and subsequently refining it to minimize travel time within given constraints. 
The trajectory search accounts for both the feasibility and the cost related to wheel movement, as well as connectivity within the configuration space. 
Consequently, we introduced the time-cost-2nd-order Hybrid-A*~\cite{hybridAstar} method for initial trajectory search, which evaluates wheel conditions and integrates travel cost into time consumption. 
In this way, the planner can search for efficient priors of the driving mode instead of manually defining them.
In the optimization process, our method referred to the Model Predictive Control~(MPC) framework, emphasizing the need for a motion state equation that is differentiable by third order and continuous to adhere to the acceleration constraints \textbf{}. 
Given that the control model by Christian \textit{et al.}~\cite{Spherical_ICM} exhibits \textbf{singularity} in steering angle and lacks differentiability when rotation radius is zero, we integrated a differentiable description of rigid body motion~\cite{murray1995differentialflat} with \textbf{steering constraints}. 
This integration was facilitated by introducing intermediary variables to circumvent the robot state singularities problem and linearizing the constraint expressions. 

Our contributions are three-fold:
\begin{enumerate}
    \item We present a time-cost-2nd-order initial trajectory planner, which is widely suitable for robots with fixed steering axis positions.
    \item We propose a new universal smoother that incorporates our kinematic model, linearizing steering constraints to generate feasible paths and control mechanisms for C-AWS.
    \item We demonstrate the digital twinning of the experimental platform and the conduction of experiments in a simulation environment verifying the effectiveness of our method.
    % 加一些定量描述？
\end{enumerate}

% Traditional stabilization purpose AWS

% Mode shift AWS

% Manual control

% Autonomous Pose-2-Pose locomotion(ours)

% Compared with traditional control papers, our method can take obstacle into consideration(not pure path following), can output wheel commands.

% Non flat model to flat model

% MINLP

% Wheel Constrain

% Contribution:

% \textbf{All-Wheel-Steering Local Planning System (TEB)}
    
% \space - High flexibility

% \space - Higher efficiency

% \textbf{Parameter Autoadaption}

% \space - Measurement and optimizing method
%     % - G2O optimizer(implementation detail)

% \textbf{unconstructed scenario (HAtar) }
    
% \space - Hybrid A Star Global planning

% \space - How to define the searching space

% \textbf{Modeling and solving}

% \space - Modeling

% \space - Constrain(avoid arctan2)

% \textbf{experiments in both simulator and field}

% \space - parking

% \space - locomotion

% Problem Formulation
% 优化变量
%% 中心点位置
%% 到四个控制量 
%% 优化目标
%% 约束量
% 轨迹规划MPC->TEB 硬约束到软约束
% dead lock problem

\section{Related Works}

\subsection{Kinematic Models}
\label{rel:model}
Kinematic models are represented as ordinary differential equations. 
These equations describe the relationship between the changing states and their rates of change over time. 
Mass point models are a simplified representation where moving states are linearly additive.
% , like when the heading of the controlled object is independent of its motion. 
% This model is suitable for holonomic moving platforms equipped with Macnumm wheels~\cite{macnum}.
% TODO remove LQR
Control problems of such models can be classified as Linear Programming~(LP) problems and can usually be solved with simple negative feedback control or minimizing linear quadratic cost functions.
% Linear Quadratic Regulator~(LQR).
Non-holonomic dynamic models are used to describe physical systems with dependent states. 
Controlling with this kind of model typically involves numerical Nonlinear Programming~(NLP) solving methods~\cite{Ma2022Alternating,10171831}.
For instance, single-wheel models are used on differential steering wheeled robots~\cite{saidonr2011differentialwheel}, which adjust the speed difference between drive wheels to change the facing direction of the vehicle and adjust its moving direction.
The bicycle model is another non-linear model that simplifies the mathematical representation of the dynamics of vehicles with Ackerman geometry. 
This model captures the essential motion features of vehicles such as bicycles, whose rear wheels cannot be steered or whose rear wheels are steered in a certain proportion~\cite{acker4ws}.
% , and even trailers~\cite{murray1993nonholonomic}. 
The steering angle has an injective relationship to the curvature of an arc tangent to the heading direction at a specific point.
In contrast to the aforementioned nonlinear dynamic models, AWS robots allow for dynamic adjustment of the rotating center.
This transforms the control of lateral movement from one-dimensional to two-dimensional, significantly enhancing the robot's mobility. 
However, in the planning and control of AWS robots, the relationship between the wheel's steering angle and the motion state is non-injective and is further complicated by substantial inter-wheel constraints. 
These issues amplify the non-linearity and discontinuity of the model, rendering it unsuitable for direct resolution through NLP solvers.
In our C-AWS algorithm, we reformulated this problem to enable its solution using NLP solvers.

% \subsection{Trajectory Continuity}

% Trajectory optimization for vehicles at each timestep utilizes parametric curves or numerical methods to maintain state continuity.
% Piecewise parametric curves, such as Dubins\cite{dubins} and Redd's Shepp\cite{reddshepp} paths, segment intricate paths into concatenated simpler maneuvers like constant turning and forwarding in straight lines while spiral curves can provide gradual changes in direction and speed\cite{nmp}.
% Splines are generated by interpolating between control points\cite{ssc} while polynomials are determined by coefficients\cite{han2023differential}. 
% However, the use of parameterized curves introduces bias related to curve property or degree of freedom.
% To solve these trajectory optimization problems numerically, methods such as Euler\cite{dontchev2001euler} and advanced Runge-Kutta (RK) algorithms are employed in our method. 
% Specifically, RK4 offers enhanced accuracy in estimating solutions to motion-governing differential equations\cite{MPCC}. 

\subsection{Trajectory Planning}
\label{rel:planning}
Trajectory planning is a process that generates a feasible and followable path for a robot, assigning specific speed and pose requirements at certain time steps. This process is governed by two core criteria: ensuring that the trajectory is collision-free and trackable~\cite{DARPA}.
Sampling-based methods are popular in robotics for finding collision-free paths due to their versatility in incorporating user-defined objectives. These methods encompass state lattice~\cite{lattice, frenet} approaches and probabilistic planners. 
They essentially operate by sampling the robot's possible states within the configuration space to discover a viable trajectory between the start and goal nodes.
Lattice-based planners break down the continuous state space into a discrete lattice graph structure for planning purposes. 
Subsequently, algorithms such as Dijkstra's algorithm are applied to identify the optimal trajectory through this graph.
Excluding lattice planners, probabilistic planners exemplified by Rapidly-exploring Random Trees (RRT)~\cite{rrt-ori} generate a feasible path by iteratively growing a tree of states starting from the initial node. 
The sampling process inherent in these methods~\cite{kinodynamicrrt,trackrrt,hybridAstar} accounts for the dynamic constraints of the vehicle. 
In the process of searching for paths, piecewise-parametric curves are extensively used. 
% For example, Dubins' paths~\cite{dubins} and Reed-Shepp's paths~\cite{reddshepp} decompose complex paths into sequences of simpler maneuvers, such as constant turning and straight-line motion. 
% Meanwhile, spiral curves facilitate gradual changes in direction and speed~\cite{nmp}. 
Specifically, the forward movement of bicycles can be modeled as an arc, ensuring feasibility for the non-holonomic model in Hybrid-A*\cite{hybridAstar}.
Therefore, our time-cost-2nd-order trajectory searching method has been developed based on the Hybrid-A* algorithm, which can efficiently take advantage of this feature.

MPC techniques address trajectory planning by conceptualizing it as an Optimal Control Problem (OCP), subsequently transformed into an NLP problem~\cite{mpc-non-holo,MPCC,mpc-carlike}. 
Regarding the functionality of MPC, our algorithm employs this framework with meticulous model design and high-quality initial guesses. 
% This is crucial for meeting complex, possibly non-convex, trajectory constraints while improving solution speed and robustness.
% This method efficiently optimizes both discrete states and control inputs, incorporating dynamic and safety constraints. 
Trajectory optimization for vehicles at each time step utilizes differentiable curves or numerical methods to maintain state continuity.
Differentiable curves, such as splines, are generated by interpolating between control points~\cite{ssc}, while polynomials are determined by coefficients~\cite{han2023differential}. 
However, the use of parameterized curves introduces bias related to the curve property or degree of freedom.
To solve these trajectory optimization problems numerically, we employ methods such as Euler~\cite{dontchev2001euler} and advanced Runge-Kutta (RK) algorithms. 
Specifically, we utilize the RK4 algorithm, which strikes a balance between accuracy and computational efficiency.
\section{Methodology}

\subsection{Rigid Body Vehicle Model}
% In \cite{murray1995differentialflat} one force with one torque is differential flat. So our control model is established accordingly.

First, we define the state variables and control variables of the system as an arbitrary rigid body 
\begin{equation*}
    \begin{split}
        \mathbf{x} &= 
        \begin{bmatrix} 
            x, y, \theta, \dot{x}, \dot{y}, \dot{\theta}
            % , \{v_{w}\}, \{\delta_{w}\} 
        \end{bmatrix} 
        ^\top, \\
        \mathbf{u} &= 
        \begin{bmatrix}
            \ddot{x}, \ddot{y}, \ddot{\theta}
        \end{bmatrix} 
        ^\top,
        % extended control variables [c_x, c_y] 
    \end{split}
\end{equation*}
where $x,y,\theta$ are the essential descriptions of 2D movement of a rigid body: position and heading angle.
The wheel positions regarding the robot frame are defined as
\begin{equation*}
    \mathbf{W} := \{\mathbf{w}_i|\mathbf{w}_i \in \mathbb{R}^2 , i \in \{1,2,\ldots,N_w\}\},
\end{equation*}
and steering limits for each wheel are defined as
\begin{equation*}
    \begin{split}
        \bar{\mathcal{D}}_{lim} &:= \{\bar{\delta}_{lim,w_i}|\bar{\delta}_{lim,w_i} \in \mathbb{R}, \text{for} \ w_i \in \mathbf{W}\}, \\
        \underaccent{\bar}{\mathcal{D}}_{lim} &:= \{\underaccent{\bar}{\delta}_{lim,w_i}|\underaccent{\bar}{\delta}_{lim,w_i} \in \mathbb{R}, \text{for} \ w_i \in \mathbf{W}\}.
    \end{split}
\end{equation*}

The pose $\mathcal{P}$ of a rigid body in 2D space can be represented by 2 dimentional Special Euclidean Group $SE(2)$, i.e.,
\begin{equation*}
\mathcal{P} = 
\begin{bmatrix}
\mathbf{R} & \mathbf{p} \\
0 & 1
\end{bmatrix},
% \quad \text{, where} \quad \mathbf{R} \in SO(2) \quad \text{and} \quad \mathbf{p} \in \mathbb{R}^2,
\end{equation*}
where the rotation $\mathbf{R}$ is belong to 2 dimentional Special Othogonal Group $SO(2)$ and position $\mathbf{p}$ is a 2 dimentional vector, i.e.,
\begin{equation*}
        \mathbf{R}(\theta) = \begin{bmatrix}
            \cos{\theta} & -\sin{\theta} \\
            \sin{\theta} & \cos{\theta}
        \end{bmatrix} \in SO(2) \text{,~}
        \mathbf{p} = \begin{bmatrix}
            x \\
            y
        \end{bmatrix} \in \mathbb{R}^2.
\end{equation*}
By deriving the rotation and position transformation regarding $t$ respectively, we can get
\begin{equation*}
    \begin{split}
        \bm{\omega}(\theta) &= \frac{\partial \mathbf{R(\theta)}}{\partial t} = \frac{\partial}{\partial t}\begin{bmatrix}
            \cos\theta & -\sin\theta \\
            \sin\theta & \cos\theta
        \end{bmatrix} \\
        &=
        \begin{bmatrix}
            -\sin\theta & -\cos\theta \\
            \cos\theta & -\sin\theta
        \end{bmatrix} \cdot \dot{\theta} \\
        % \frac{d\theta}{dt}
        &= \mathcal{K}(\theta)\dot{\theta}, \\
        \bm{v} &= \frac{\partial \mathbf{p}}{\partial t} = \begin{bmatrix}
            \dot{x} \\
            \dot{y}
        \end{bmatrix}.
    \end{split}
\end{equation*}

\begin{figure}[!t]
\centering
\includegraphics[width=0.8\linewidth]{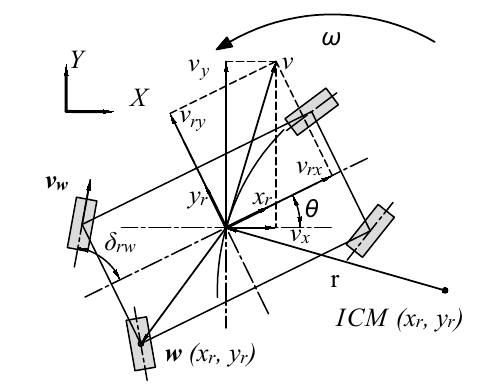}
\caption{The schematic representation of an AWS robot kinematic model. Note that the $\omega$ direction only indicates the positive direction, instead of the rotating direction at the instance.}
\label{fig:annotation}
\vspace{-1.5em}
\end{figure}

Since the instantaneous motion of the vehicle can be regarded as a rotation around a certain point, in order to calculate the motion of each point on the vehicle body, we obtain the position of the instantaneous center which is indicated by the radius vector 
\begin{equation}
    \mathbf{r}_{ICM} =
    \bm{\omega}^{-1}(\theta)\bm{v}
    =
    \mathcal{K}(\theta)^\top 
    \begin{bmatrix}
        \frac{\dot{x}}{\dot{\theta}} \\
        \frac{\dot{y}}{\dot{\theta}}
    \end{bmatrix},
    \label{eq:wvr}
\end{equation}
where $ \mathbf{r}_{ICM} $ stands for the vector point from the relative center of the ICM position pointing to the vehicle control center \textbf{in the robot frame}. 
Here we introduced the wheel moving velocity as an intermediate variable.
For each wheel ground contacting position, the required moving velocity in the world frame can be computed by
\begin{equation*}
    \begin{split}
    \mathbf{r}_{ICM,w} &= \mathbf{r}_{ICM} + \mathbf{w}, \\
    \bm{v}_{w}
    &= 
    \bm{\omega}(\theta)\cdot\mathbf{r}_{ICM,w}
    % = \mathcal{R}(\theta)\cdot\dot{\theta}\cdot(\mathbf{r}_r+\mathbf{w})
    = \bm{v}+\bm{\omega}(\theta)\mathbf{w},
    % = \\
    % \begin{bmatrix}
    %     v_{yrw} \\
    %     v_{xrw}
    % \end{bmatrix}
    % &=
    % \begin{bmatrix}
    %     \omega (x_{rw}-r\cos{\phi}) \\
    %     \omega (y_{rw}-r\sin{\phi})
    % \end{bmatrix}
    \end{split}
\end{equation*}
while the wheel moving speed regarding robot heading is
\begin{equation}
    \bm{v}_{r,w} = \mathbf{R}(\theta)^{-1}\bm{v}_{w} = \mathbf{R}(\theta)^{\top}\bm{v}_{w},
    \label{eq:rel_v}
\end{equation}
% \begin{equation*}
%         \begin{split}
        % &= 
        % \arccos{
        % \frac{
        % \bm{v}_{w} \cdot \mathbf{R}(\theta_w+\theta) \mathbf{i}_{rx}
        % }{
        % |\bm{v_{w}}|\cdot 1
        % }
        % }\\
%         \end{split}
% \end{equation*}
%%%%%%%%%%%%%%%%%%%%%%%%%%%%%%%%%%%%%%%
the steering angle can be written as
\begin{equation}
\label{eq: steer}
    \bm{\delta}_{w} 
    = \arctan_2{(v_{wy},v_{wx})} %- (\theta_w + \theta)
    = 2\arctan \left(\frac{v_{wy}}{|\bm{v}_w|+v_{wx}}\right),
\end{equation}
and in the robot frame, it would be
\begin{equation}
     \bm{\delta}_{r,w} = \bm{\delta}_{w} - \theta.
     \label{eq:rel_steering}
\end{equation}
The time derivative of \( \bm{v}_{w} \) is
\begin{equation}
    \label{eq:diff_vw}
    \begin{split}
        \frac{\partial \bm{v}_{w}}{\partial t} 
        &= \frac{\partial \bm{v}}{\partial t} + \frac{\partial \mathbf{\omega}(\theta)}{\partial t} \mathbf{w} \\
        &=\begin{bmatrix}
            \ddot{x} \\
            \ddot{y}
        \end{bmatrix}
        + \begin{bmatrix}
            -\cos{\theta} & \sin{\theta} \\
            -\sin{\theta} & -\cos{\theta}
        \end{bmatrix} \dot{\theta}^2 \mathbf{w} \\
        &+ \begin{bmatrix}
            -\sin(\theta) & -\cos(\theta) \\
            \cos(\theta) & -\sin(\theta)
        \end{bmatrix}\ddot{\theta}\mathbf{w}.
        % &= \begin{bmatrix}
            % \ddot{x} - \dot{\theta}^2 w_x - \ddot{\theta}w_y \\
            % \ddot{y} - \dot{\theta}^2 w_y + \ddot{\theta}w_x
            % \end{bmatrix}
    \end{split}
\end{equation}
%%%%%%%%%%%%%%%%%%%%%%%%%%%%%%%%%%%%%%%
% and time derivative of \( \delta_{w} \) is,
% % % https://en.wikipedia.org/wiki/Atan2#:~:text=domain%20in%20question.-,Derivative,-%5Bedit%5D
% \begin{equation}
% \label{eq:diff_steer}
%     \begin{split}
%         \frac{\partial \bm{\delta}_{w}}{\partial t} 
%         &= \frac{\partial \arctan_2{(v_{wy},v_{wx})}}{\partial t} \\%- \frac{\partial (\theta_w + \theta)}{\partial t} \\
%         % &= \frac{v_{wx}\dot{v}_{wy}-v_{wy}\dot{v}_{wx}}
%         % {
%         %     |\mathbf{v}_w|^2
%         %     % v_{wx}^2+v_{wy}^2
%         % } - \dot{\theta}
%         &= \frac{-v_{wy}\dot{v}_{wx}v_w+v_{wx}\dot{v}_{wy}v_w-v_{wx}v_{wy}-v_{wy}^2+v_w^2\dot{v}_{wy}}{2v_w^2(v_w+v_{wx})}.
%     \end{split}
% \end{equation}
% % Nevertheless, it is not used in the optimizing process together with eq.~\ref{eq: steer}, eq.~\ref{eq:rel_steering} to avoid discontinuity and singularity.

\subsection{Path Initialization}
\begin{figure}[t]
  \centering
  % \begin{subfloat}
    % \centering
    \vspace{0.4em}
    \hspace{-0.5em}
    \includegraphics[width=0.34\linewidth]{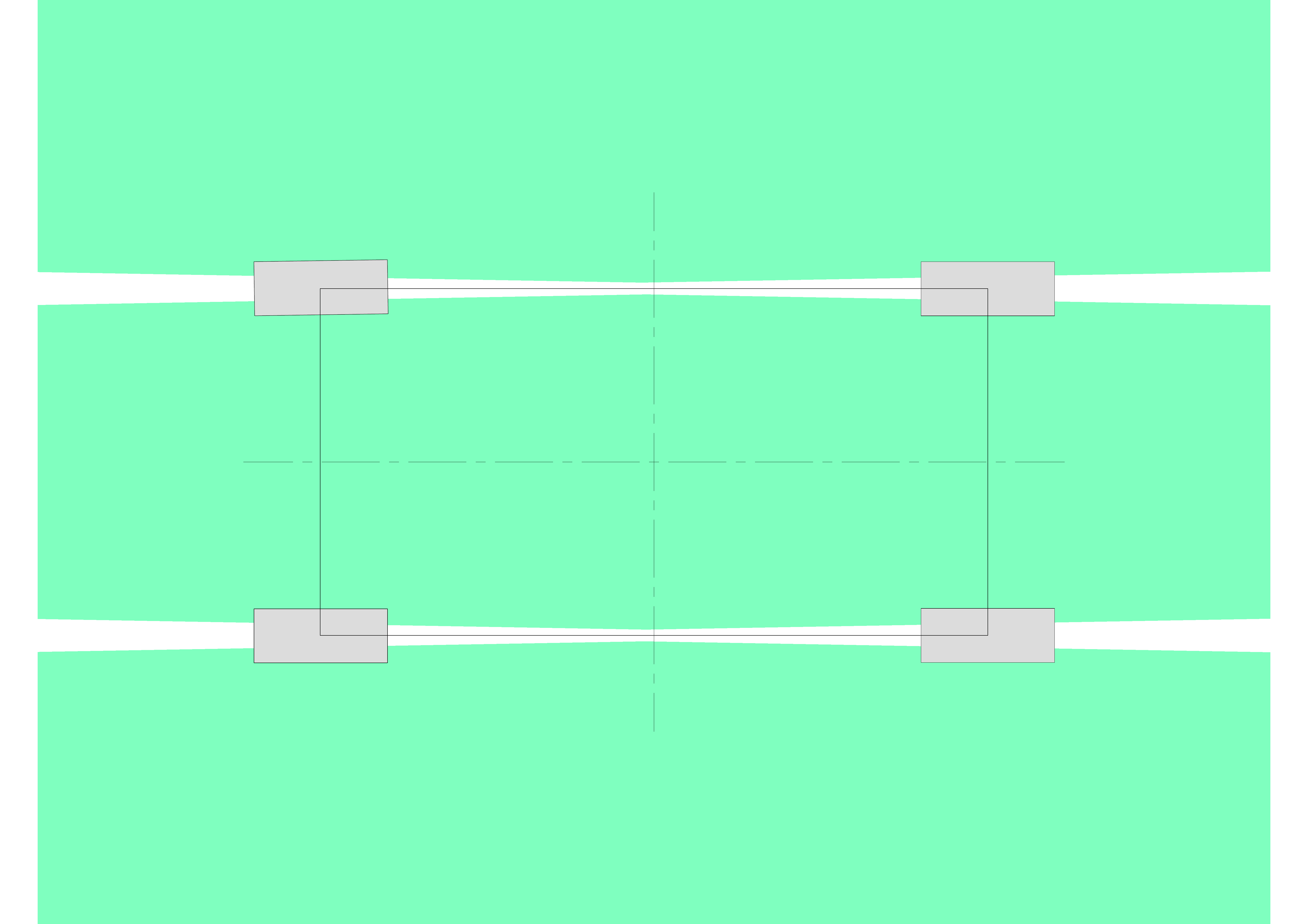} \hspace{-0.8em}
    \includegraphics[width=0.34\linewidth]{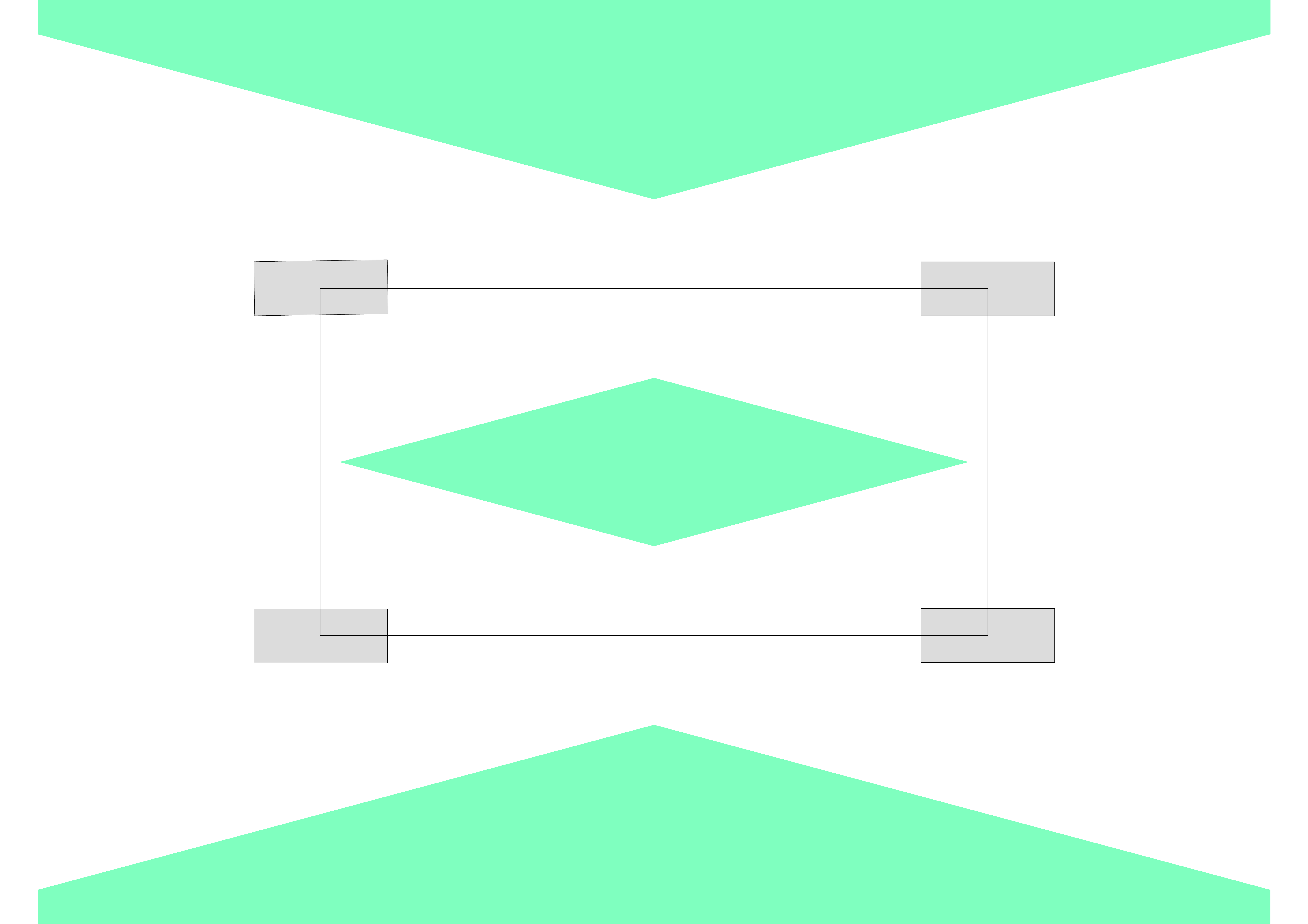} \hspace{-0.8em}
    \includegraphics[width=0.34\linewidth]{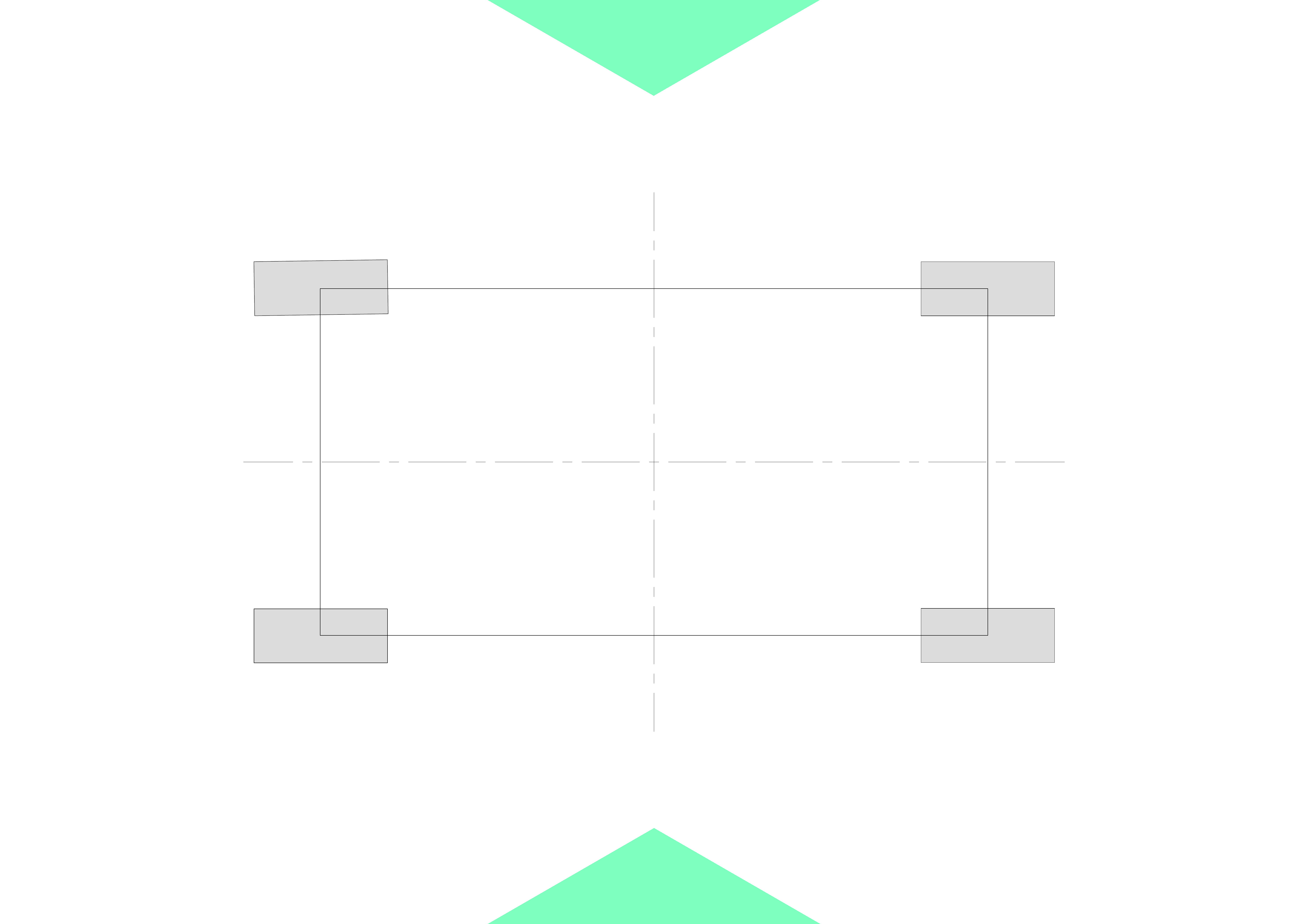} \hspace{-0.8em}
    \caption{Feasible ICM regions near control center are marked green, here we show feasible regions when  $\delta_{lim}=$ 90\textdegree, 75\textdegree, 60\textdegree.}
    \label{fig:sampling_region}
    % \vspace{-1em}
  % \end{subfloat}
\end{figure}
\begin{figure}[t]
    \centering
    % \hspace{-1.5em}
    \includegraphics[width=0.5\linewidth]{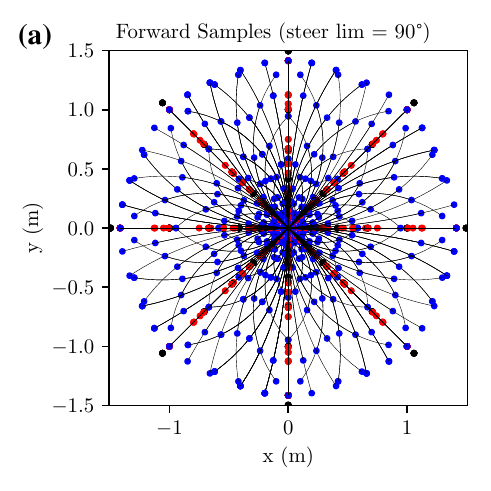} \hspace{-0.5em}
    \includegraphics[width=0.5\linewidth]{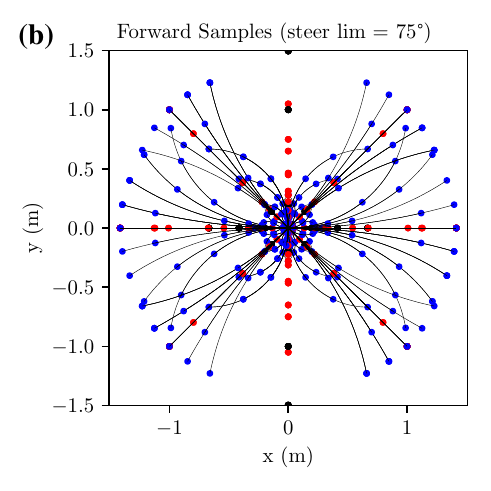}\\
    \vspace{-0.5em}
    \hspace{-1.5em}
    \includegraphics[width=0.5\linewidth]{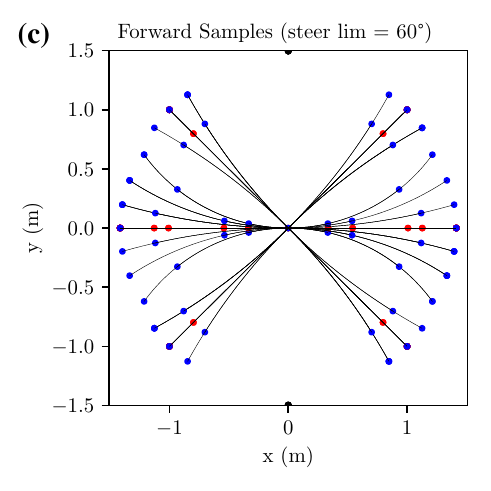} \hspace{-0.5em}
    \includegraphics[width=0.5\linewidth]{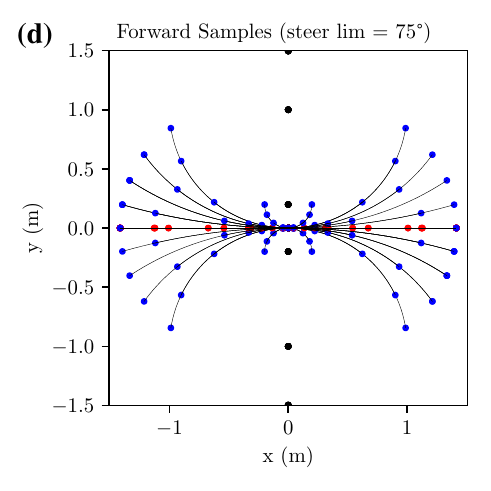}
    \caption{The sampled maneuvers with $\delta_{lim}=$ 90\textdegree, 75\textdegree, 60\textdegree~and front-75\textdegree-rear-0\textdegree~for (a), (b), (c) and (d) respectively. The red points are the end of the velocity vector, the blue points are the end of trajectories, and the black points are instantaneous centers. In all the forward samples above, $N$ of $\mathcal{E}$ is 8, $N$ of $\Psi$ is 8, and $\omega$ is uniformly sampled 8 times in range $[-\pi/2, \pi/2]$.}
    \label{fig:sampled_maneuver}
    \vspace{-1.6em}
\end{figure}

% \vspace{-0.8em}
To offer a high-quality feasible route for the optimizer, a global planning method based on Hybrid-A* is developed.
This will reduce the failure of the NLP solver and increase the solving speed.
In our design, not only the first-order states like position and heading are sampled, but also the body velocity is sampled and recorded.
We simplify the integration process from high-order variables to low-order variables, which is a trade-off in calculation time and accuracy.
For the forward simulation process, we customized the grid sampling method for this all-wheel steering chassis.
To encourage that each sampling explores new states, we conduct grid sampling of ICM on the spherical coordinate to deal with singularity when steering is zero and align with the sequential optimization problem. 
% The condition that the ICM position needs to meet is that it must be within the intersection of the areas that the normals of the movement directions of all wheels can pass through.
The feasible ICM location regarding steering limits is shown in Fig.~\ref{fig:sampling_region}. 
It must be within the intersection of the areas that the normals of the movement directions of all wheels can pass through.

% $\Delta x$: position sampling values, which are advised to be in the same magnitude or times the map resolution.

% $\theta$: yaw resolution, which is advised to be in the same magnitude or times the yaw resolution.

% \begin{equation*}
%     \begin{split}
%         \Delta x &= \omega\times \mathbf{r}\\
%         or \\ 
%         \Delta \theta &= \frac{\mathbf{v}}{\mathbf{r}}
%     \end{split}
% \end{equation*}
The radius vector $\mathbf{r}$ points from ICM to the vehicle control center is used here again to represent the moving states.
% which also stands for the wheel maneuver of the vehicle. 
% The sampling region is determined by the vehicle's dynamic feasible region.
Two variables $\varepsilon$ and $\psi$ are uniformly sampled variables on the coordinate axes of the spherical coordinate system, which can be written as
\begin{equation*}    
\begin{split}
\mathcal{E} : \left\{ \varepsilon_i| \varepsilon_i = \frac{\pi i +\epsilon}{2N+\epsilon}, \forall i \in \left\{0,1,\ldots,N\right\} \right\},\\
\Psi : \left\{ \psi_i| \psi_i = -\pi+\frac{2\pi i}{N},\forall i \in \left\{0,1,\ldots,N\right\} \right\}.
\end{split}
\end{equation*} 
On the grid mesh points constructed by the two sets, we get $\mathbf{r}$ samples by
\begin{equation*}
    \begin{split}
        \mathbf{r} &= 
        \begin{bmatrix}
            -\tan{\varepsilon} \cdot \cos{\psi} \\
            -\tan{\varepsilon} \cdot \sin{\psi}
        \end{bmatrix}, \forall \varepsilon \in \mathcal{E}, \forall \psi \in \Psi, \\
        \text{subject to}, \\
        % \mathbf{r}_{ICM,w} &\times 
        % \begin{bmatrix}
        %     \cos{(\delta_{ulim}+\frac{\pi}{2})} \\
        %     \sin{(\delta_{ulim}+\frac{\pi}{2})}
        % \end{bmatrix}
        % \cdot \mathbf{r}_{ICM,w} \times
        % \begin{bmatrix}
        %     \cos{(\delta_{llim}+\frac{\pi}{2})} \\
        %     \sin{(\delta_{llim}+\frac{\pi}{2})}
        % \end{bmatrix} \leq 0, \\
        &\underaccent{\bar}{\delta}_{w} \leq |\arctan_2(\mathbf{r}_{y,ICM,w},\mathbf{r}_{x,ICM,w})|-\frac{\pi}{2} \leq \bar{\delta}_{w} \\
        &\mathbf{r}_{ICM,w}, 
        \bar{\delta}_{w} \in \mathcal{D}_{ulim}, \underaccent{\bar}{\delta}_{w} \in \mathcal{D}_{llim},
         \forall w \in \mathbf{W}.\\
    \end{split}
\end{equation*}
The rotation speed $\omega$ can be directly sampled in a certain range.
However, as $r$ increases, certain rotation angular regarding that time step may cause the traveling distance to be too long, so we set the longest curvature length to control the farthest traveling distance of each time step. 
% The sampled $\omega$ can be confined by a maximum curvature length.
% \[\omega = \max (\frac{L_{curve}}{|\mathbf{r}|},\omega_{sample}).\]
The sampling results of different steering limits are demonstrated in Fig.~\ref{fig:sampled_maneuver}.
Through simple mathematical derivation according to Eq.~(\ref{eq:wvr}), we can obtain the current speed $\Delta v$ and next position $\Delta x$ after a certain time under this r and omega. 
Also, we can derive the relative wheel speed $v_{w,n}$ and steering angle $\delta{w,n}$ of that node according to Eq.~(\ref{eq:rel_v}), and Eq.~(\ref{eq:rel_steering}).
It is worth noting that for the common situation where the limit angle is less than 90\textdegree, there is a single mapping relationship between the wheel angle speed and the wheel friction point speed. 
If it exceeds 90\textdegree, an additional mapping table will be needed. 
We will not discuss this for the time being.
% complex situations that require additional definitions.
We add the speed and position values as attributes of the next point to the new node constructed during the Hybrid-A* search.
In order to better utilize the speed attribute of the node, we express the cost in time to unify the cost function and avoid weight adjustment of speed and distance information.
% \begin{figure}
%     \centering
%     \includegraphics[width=8.3cm]{img/sampled_maneuver.jpg}
%     \caption{The Sampled Maneuver for HubridA\star.}
%     \label{fig:sampled_maneuver}
% \end{figure}
So the traveling cost is computed by,
\begin{equation*}
    \begin{split}
        {t}_{v_w} &= \frac{\max(|{v}_{w,n}-{v}_{w,n-1}|)}{\dot{v}_{w,max}}, \\
        {t}_{\delta_w} &= \frac{\max(|{\delta}_{w,n}-{\delta}_{w,n-1}|)}{\dot{\delta}_{w,max}}, \\
        {t}_w &= \sqrt{k_{v_w}{t}_{v_w}^2+k_{\delta_w}{t}_{\delta_w}^2},  k_{v_w},k_{\delta_w} \in \mathbb{R},\\
        t_{body} &= \max{(\frac{\Delta x}{v_{max}}, \frac{\Delta \theta}{\dot{\theta}_{ max}})}, \\
        \Delta g &:= \max{({t}_w, t_{body})}, \forall w \in \mathbf{W}.
    \end{split}   
\end{equation*}
The heuristic term is proportional to the Euler distance to the goal, which is calculated by
\begin{equation*}
    \begin{split}
        \hat{t}_v &= \frac{\sqrt{(x_{goal}-x)^2+(y_{goal}-y)^2}}{v_{max}}, \\
        \hat{t}_{\theta} &= \frac{(\theta_{goal}-\theta)}{\dot{\theta}_{max}},  \\
        h &:= k_h\cdot\max{(\hat{t}_v, \hat{t}_{\theta})}, k_h \in \mathbb{R},
    \end{split}
\end{equation*}
where $ k_h $ is the hypothesis coefficient, which represents the ratio of the time consumption of locomotion to the time consumption of the vehicle's motion with the maneuver.
The algorithm is the same as ordinary Hybrid-A* in the unmentioned aspects.
The discretization search process of initial guess will also output the wheel steering and vehicle discontinuous state space flags, which will be explained in the next section.
% Other aspects of the algorithm follow the general rules of the Hybrid A* algorithm as introduced in \cite{Kurzer1057261}.

% Each node saves the 0-order and 1-order derivatives of the vehicle's state, which are used to calculate the time consumption of the maneuver.

\subsection{Trajectory Optimization}
%规避车轮的奇点，创新性地使用车轮速度作为中间变量，提高模型的线性化程度
% The slack variable is convex.
% Dynamic constraints:
% RK4 to make the system approximately continuous.
% The fourth-order Runge-Kutta method is an iterative algorithm that estimates the next value of a function at a series of discrete time steps based on the derivative of the function at various points within that step. 
% This technique provides an approximation with an error term that is proportional to the square of the step size.
We choose the fourth-order Runge-Kutta method  to ensure continuity between states at different times within time horizon $T$, as shown by the formula:
\begin{equation*}
\mathbf{x}(t + dt) = \mathbf{x}(t) + \frac{1}{6} (k_1 + 2k_2 + 2k_3 + k_4),
\end{equation*}
% where,
% \begin{align*}
% k_1 &= dt_t \cdot f(t, y(t)) \\
% k_2 &= dt_t \cdot f\left(t + \frac{dt_t}{2}, y(t) + \frac{k_1}{2}\right) \\
% k_3 &= dt_t \cdot f\left(t + \frac{dt_t}{2}, y(t) + \frac{k_2}{2}\right) \\
% k_4 &= dt_t \cdot f\left(t + dt_t, y(t) + k_3\right)
% \end{align*}
where
\( \mathbf{x}(t) \) represents the states at time \( t \),
\( dt \) is a short time duration,
% \( f(t, y) \) is the function defining the differential equation \( \frac{\partial y}{\partial t_t} = f(t, y) \),
\(k_1,k_2,k_3,k_4\) are estimated slopes by the ordinary differential equation \( \frac{\partial \mathbf{x}}{\partial t} = \dot{\mathbf{x}}=\begin{bmatrix}
            &\dot{x}
            &\dot{y}
            &\dot{\theta}
            &\mathbf{u}^\top
            % &\frac{\partial \mathbf{v}_{w}}{\partial t}^\top
            % &\frac{\partial \bm{\delta}_{w}}{\partial t}^\top
        \end{bmatrix}^\top \), by this case, the continuous constrain between \(\mathbf{x}(t)\) and \(\mathbf{x}(t+dt)\) is
% \begin{equation*}
    % \begin{split}
        % f(t, y) = \dot{\mathbf{x}}
        % \begin{bmatrix}
        %     \dot{x} 
        % &\dot{y} 
        % &\dot{\theta} 
        % &\ddot{x} 
        % &\ddot{y} 
        % &\ddot{\theta} 
        % &\dot{v}_{wheels} 
        % &\dot{\delta}_{wheels}
        % \end{bmatrix}^\top \\
        % =
        % \begin{bmatrix}
        %     &\dot{x}
        %     &\dot{y}
        %     &\dot{\theta}
        %     &\mathbf{u}^\top
            % &\frac{\partial \mathbf{v}_{w}}{\partial t}^\top
            % &\frac{\partial \bm{\delta}_{w}}{\partial t}^\top
        % \end{bmatrix}^\top
    % \end{split}    
% \end{equation*}
\[ G_c:=\mathbf{x}(t+dt) = RK4(\mathbf{x}(t),u(t),dt), \forall t \in [0, T]. \]

% We mentioned before that there is a singular point when the wheel speed $v_w = 0$, so we first set this constraint to avoid variables falling into the singular point during the solution process. We make sure $\mathbf{v}_w \ne 0$ by
% \[G_{v_w}:=\mathbf{v}_w^2>0 \]

% 现在问题在于：
    % 1. 不知道条件能否在多个中进行选择（0*NaN还是会报错）

% \begin{equation*}
%     \begin{split}
%         \mathbbm{1}_{\delta,w} &\subseteq \{-1,1\} \\
%         % \mathbbm{1}_{v,w} &\subseteq \{-1,1\}
%     \end{split}
% \end{equation*}

% Constraints:
We simplify the calculation of wheel angle by calculating the wheel's speed vector and rotation angle limit vector. \textbf{Steering limit constraints} can be modeled by
\begin{equation*}
    \begin{split}
        % \delta_{wheels} \in [-\delta_{lim}, \delta_{lim}], \\
        G_{\delta_{lim}} &:= 
        \mathbf{v}_{r,w} \times %\mathbbm{1}_{\delta, w} \cdot
        \begin{bmatrix}
            \cos{(\bar{\delta}_{lim})}  \\
            \sin{(\bar{\delta}_{lim})}
        \end{bmatrix} 
        \cdot 
        \mathbf{v}_{r,w} \times %\mathbbm{1}_{\delta, w} \cdot
        \begin{bmatrix}
            \cos{(\underaccent{\bar}{\delta}_{lim})}  \\
            \sin{(\underaccent{\bar}{\delta}_{lim})}
        \end{bmatrix} \leq 0, \\
        & \forall t \in [0, T], \forall w \in \mathbf{W}.
    \end{split}
\end{equation*}
When the steering range is larger than 180, this term can be removed.

\textbf{Mode changing constraints:}
% For the case where the rotation angle constraint is less than 90 degrees, we can see that the position of the ICM is not continuous. 
% We have two methods to deal with this.
The control continuous space is divided into phases.
The wheel rotation directions are provided by the initial guess.
The rotating directions $D_{w} \in \{-1,1\}$ are searched according to the mapping table, minimum cost directions are selected if the steering ranges are overlapped.
The determination of the direction of this round of rotation can be modeled as a Mixed Integer Nonlinear Programing~(MINLP)~\cite{minlp_review} problem, but its solution time is very long.
The Hybrid-A* process reduced the online optimization burden that may be introduced by MINLP.

In order to avoid the discontinuity problem caused by the atan2 function near the boundary of its value range, we use the vector angle method to measure the steering size between the two frames before and after. 
% (which should be an analytic solution between $\partial \delta$ and u)
\begin{equation*}
    \begin{split}
        % &\dot{\delta}_{wheels} \in [-\dot{\delta}_{lim}, \dot{\delta}_{lim}], \\
        G_{\dot{\delta}_{lim}} &:=
        D_{w,t}\mathbf{v}_{r,w,t} \cdot D_{w,t-1}\mathbf{v}_{r,w,t-1} \\
        &- |\mathbf{v}_{r,w,t}|\cdot|\mathbf{v}_{r,w,t-1}|*\cos(\dot{\delta}_{lim}*dt) \geq 0.
        % &-\mathbf{r}_{r,w} \times %\mathbbm{1}_{\delta, w} \cdot
    \end{split}
\end{equation*}

In addition to this, due to the discontinuity in the speed state space when switching wheel modes. 
Hybrid-A* also provides an indicator for manipulation phase $M \in \mathbb{N}^{H}$ transition keyframes. 
% Only positional continuity is preserved when traveling to keyframes.
The first and last frames of different modes are set as keyframes, and the speed of this frame is 0, which can be expressed by the equation as
\begin{equation*}
    G_M := v_{t}\cdot\left(|M(t)-M(t+dt)|+|M(t)-M(t-dt)|\right)=0.
\end{equation*}

% The main problem locates in the discontinuity of arctan2
% $r = v/\omega$

% https://blog.csdn.net/keng_s/article/details/75332011

% \begin{equation*}
%     \begin{split}
%         \frac{\partial \mathbf{v}_{wr}}{\partial t}
%         &= \bm{\omega}(0)(\frac{\partial \bm{\omega}^{-1}(\theta)}{\partial t}\mathbf{v}
%             +\bm{\omega}^{-1}(\theta)\frac{\partial \mathbf{v}}{\partial t}) \\ 
%     \end{split}
% \end{equation*}

To optimize total time and release unfeasible time constraints in the initial guess, $\mathbf{dt}$ set is also optimized.
\begin{equation*}
    \begin{split}
        G_t :=\mathbf{dt}\{dt| dt \in \mathbb{R}^+, dt \in [\underaccent{\bar}{dt}, \bar{dt}]\}.\\
        % J_t = k_t\cdot\sum_{t=0}^{H} dt_t^2
    \end{split}
\end{equation*}

The objective function is to minimize the total time and the control input.
\begin{equation*}
    \begin{split}
        \min J = \int_0^T F(\mathbf{x},t)dt + \int_0^{T} \mathbf{u}(t)^\top A \mathbf{u}(t)dt + T,
    \end{split}
\end{equation*}
in discrete mode
\begin{equation*}
    \begin{split}
        \min J = \sum_{h=0}^H F(\mathbf{x},h)+ \sum_{h=0}^{H} \mathbf{u}(h)^\top A \mathbf{u}(h)dt(h) + \sum_{h=0}^{H}dt(h),
    \end{split}
\end{equation*}
s.t.
\[ \mathbf{x}(0) = \hat{\mathbf{x}}_0, \, \mathbf{x}(H) = \hat{\mathbf{x}}_f, \]
\[ G_c, G_{v_w}, G_{\delta_{lim}}, G_{\dot{\delta}_{lim}}, G_M, G_t,\]
where $A$ is the diagonal coefficient matrix, $T$ is the total time, $\mathbf{u}(t)$ is the state vector, $\bar{\mathbf{x}}_0$ and $\bar{\mathbf{x}}_f$ are the initial and final states,
$H$ is the sampling steps in the time horizon, and $k_t$ is the weight of time optimization,
$F$ is an arbitrary task-related cost function.
% As to optimization configuration, the task function aims to minimize 
% % $|d|$, where $d$ represents 
% the distance between the actual and desired vehicle trajectories.
Here we define the $L2$ distance between poses of initial guess and smoothed trajectory as the task-oriented cost function.
Besides, \textbf{wheel acceleration} can be constrained with Eq.~(\ref{eq:diff_vw}) itself.
% , avoiding complex computation in Eq.~(\ref{eq:diff_steer}).

% \input{subsections/methodology}
\section{Experiments}

% \subsection{Experiments Setup}
\subsection{Constrained Trajectory Optimization}

% CasADi \cite{CasADi} auto-differential equations

% IPOPT \cite{ipopt}

% Platform Intel i9 12900K$^\copyright$

% Simulator IssacSim \cite{issac}

% Task Function is $argmin(|d|)$
% 有时间再做
% \subsection{Cost Unified Nonholonomic HybridA*} 

% Compare with regular zero order sampling strategy, limited search nodes

% Randomly Sample 100 Goal and Start Pairs

%  Compare success rate

%  Explored points

%  AVE Min distance to Goal

%  Time To Optimize

%  Optimization Success Rate
In experiments, we employed the open-source software CasADi~\cite{CasADi} to facilitate the automatic differentiation of equations. 
% This powerful tool was instrumental in linearizing the model and devising the control strategy. 
For numerical optimization, we utilized IPOPT~\cite{ipopt}, a widely adopted and efficient nonlinear programming solver, to solve the optimization problems posed by our research. 
The experiments were conducted on a hardware platform featuring an Intel i9 12900K processor.

% By integrating these software tools and hardware resources into our experimental framework, our study was able to carry out comprehensive and rigorous testing of the theoretical models in highly realistic simulated environments, thereby ensuring the accuracy and practical relevance of our findings.

\begin{figure}[t]
    \centering
    \vspace{0.4em}
    \includegraphics[width=0.48\linewidth]{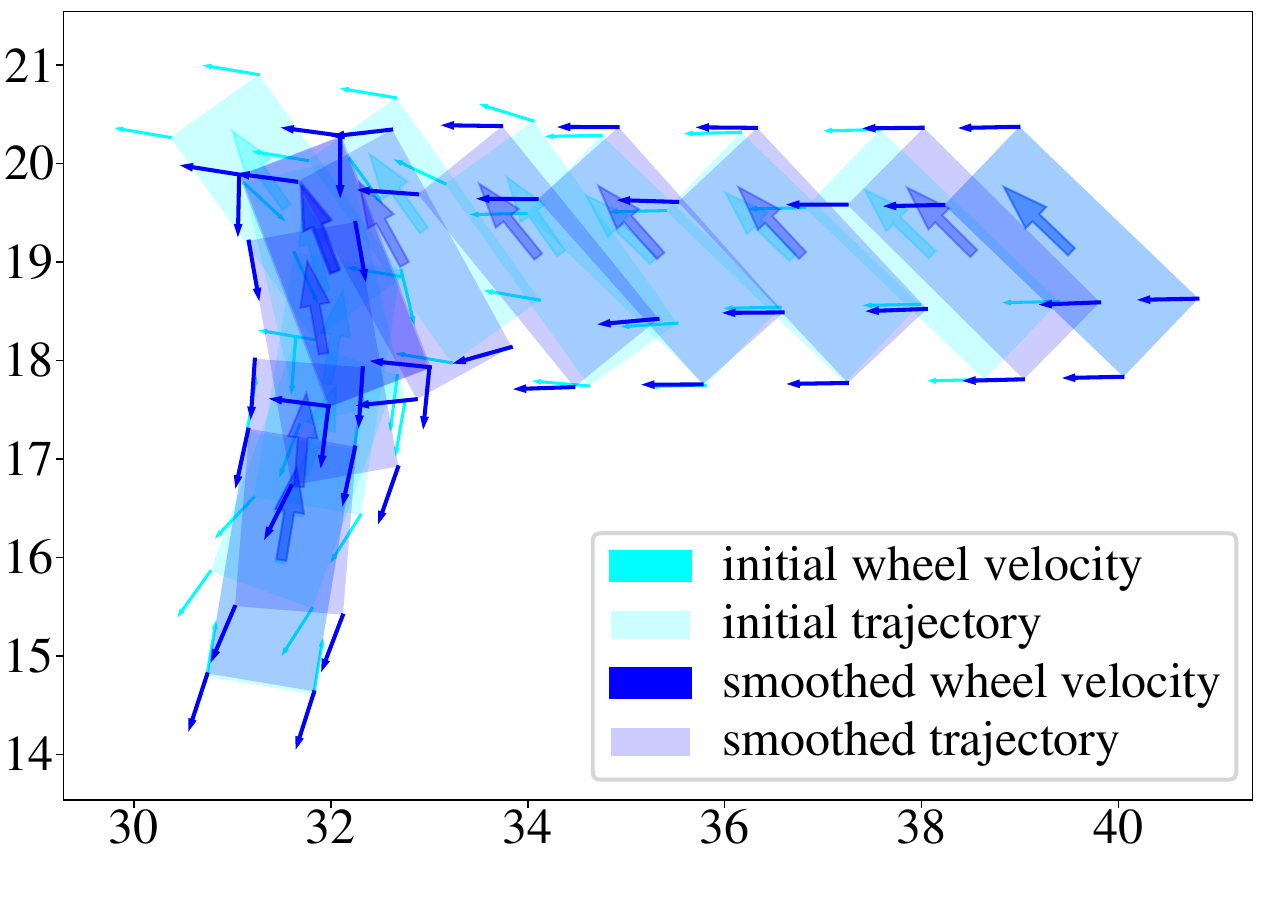}
    \includegraphics[width=0.48\linewidth]{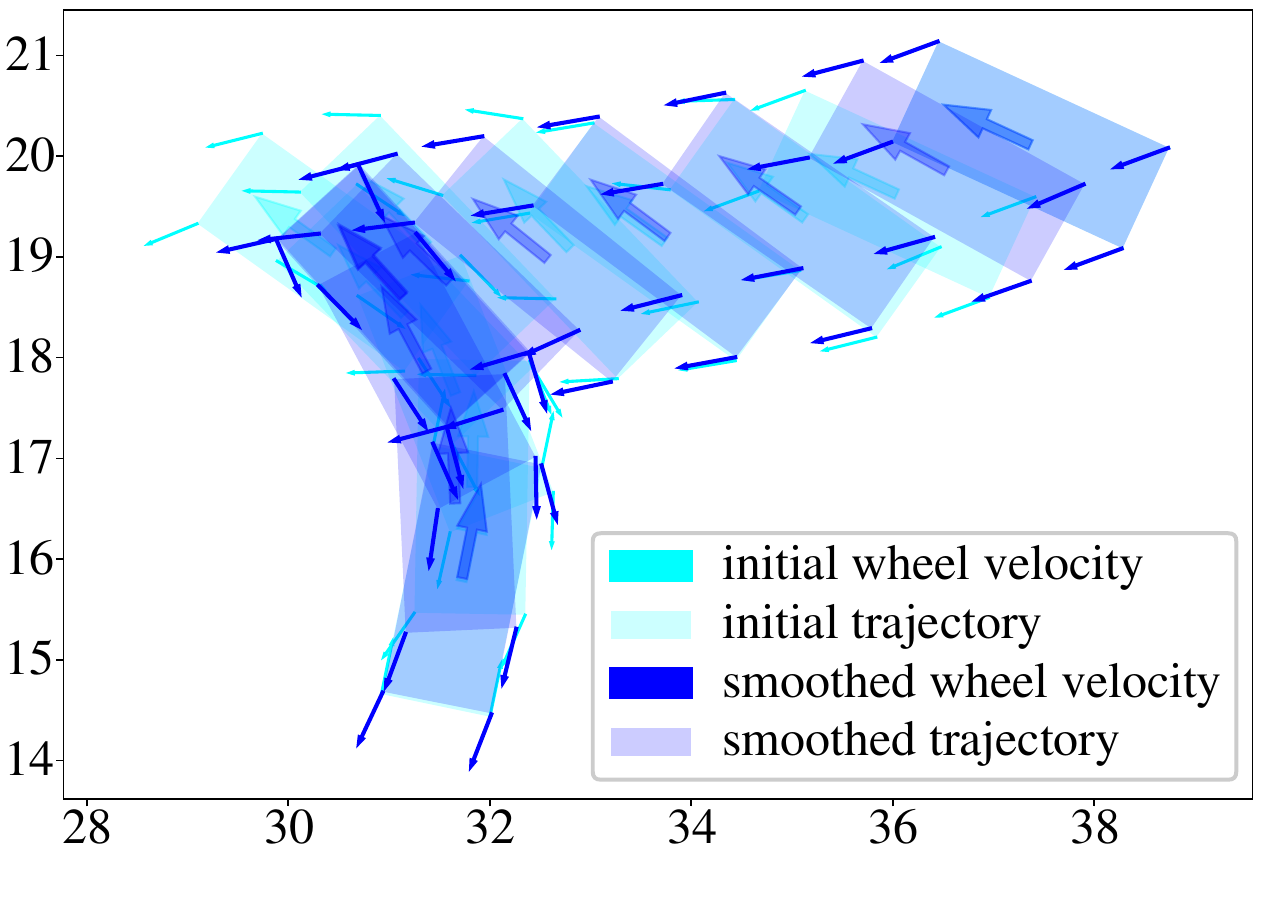}
    \\
    \vspace{-0.4em}
    % \includegraphics[width=0.9\linewidth]{img/90 deg/AWSGlobalMassPointOptimizer smoother_5.png}
    % \includegraphics[width=0.4\linewidth]{img/90 deg/AWSGlobalMassPointOptimizer smoother_3.png}
    % \includegraphics[width=0.4\linewidth]{img/75 deg/AWSGlobalMassPointOptimizer smoother_3.png}
    % \\
    % \includegraphics[width=0.5\linewidth]{img/90_deg/img_preprocess.pdf}
    \includegraphics[width=0.48\linewidth]{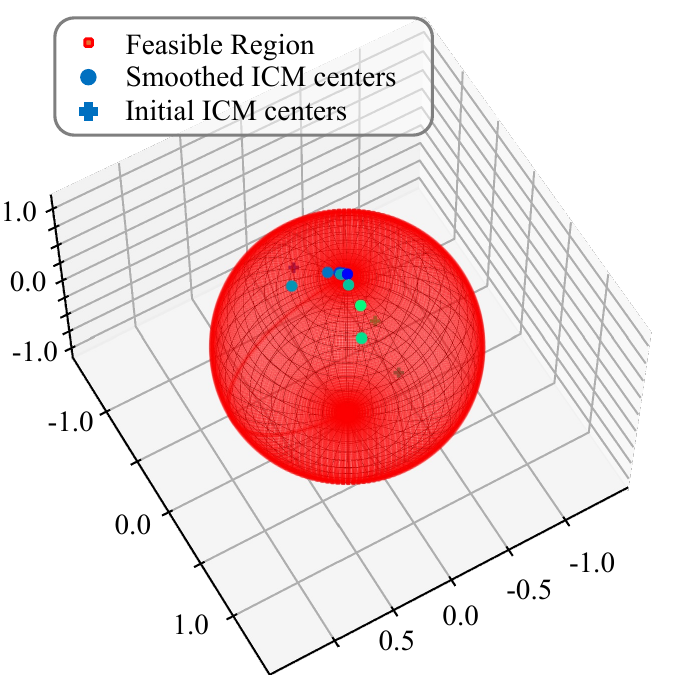}
    \includegraphics[width=0.48\linewidth]{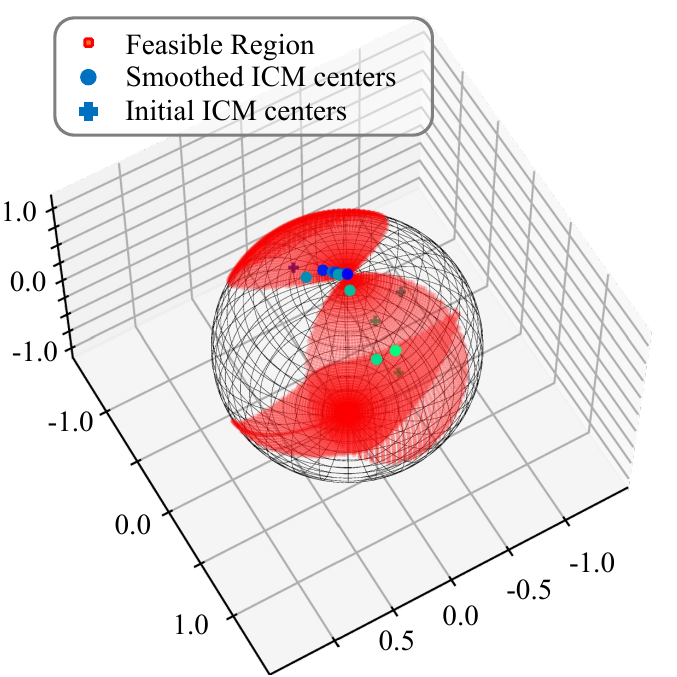}
    \caption{
    % For both Fig.~\ref{fig:instance_90} and Fig.~\ref{fig:instance_0_half}, in upper figures, the initial and refined trajectory sequences are represented in \textcolor[RGB]{51,204,204}{cyan} and \textcolor[RGB]{0,0,255}{blue}, the arrow indicating the direction of wheel speeds. 
    % \textbf{For both Fig.~\ref{fig:instance_90} and Fig.~\ref{fig:instance_0_half}:} 
    In \textbf{upper} figures, the initial and refined trajectory sequences are represented in cyan and blue, the arrow indicating the direction of wheel speeds. 
    % In lower figures, plus$(\textbf{+})$ indicate the ICM relative locations before optimization in the spherical coordinate system~\cite{Spherical_ICM}, while bullets$(\bullet)$ represent post-optimization positions, and the \textcolor[RGB]{255,0,0}{red} zone indicates the feasible solution of ICMs. 
    In \textbf{lower} figures, plus$(\textbf{+})$ indicate the ICM relative locations before optimization in the spherical coordinate system~\cite{Spherical_ICM}, while bullets$(\bullet)$ represent post-optimization positions, and the red zone indicates the feasible solution of ICMs. 
    % The color of the dot gradually transitions from \textcolor[RGB]{0,0,204}{blue} to \textcolor[RGB]{0,204,128}{green} as time progresses. 
    The color of the dot gradually transitions from blue to green as time progresses. 
    \textbf{In this figure:} front and rear steering limits are set to $\pm90$\textdegree~and $\pm75$\textdegree~in the left and right columns, respectively.
    }
    \label{fig:instance_90}
    \vspace{-1.5em}
\end{figure}
% \begin{figure}[!t]
%     \centering
%     % \includegraphics[width=0.9\linewidth]{img/AWSGlobalMassPointOptimizer smoother.pdf}
%     % \includegraphics[width=0.9\linewidth]{img/75 deg/AWSGlobalMassPointOptimizer smoother_5.png}
%     \caption{An instance of smoothed trajectory. Both front and rear steering limits are set to $\pm75$\textdegree .}
%     \label{fig:instance_75}
%     \vspace{-1.5em}
% \end{figure}
\begin{figure}[h]
    \centering
    \vspace{0.4em}
    \includegraphics[width=0.48\linewidth]{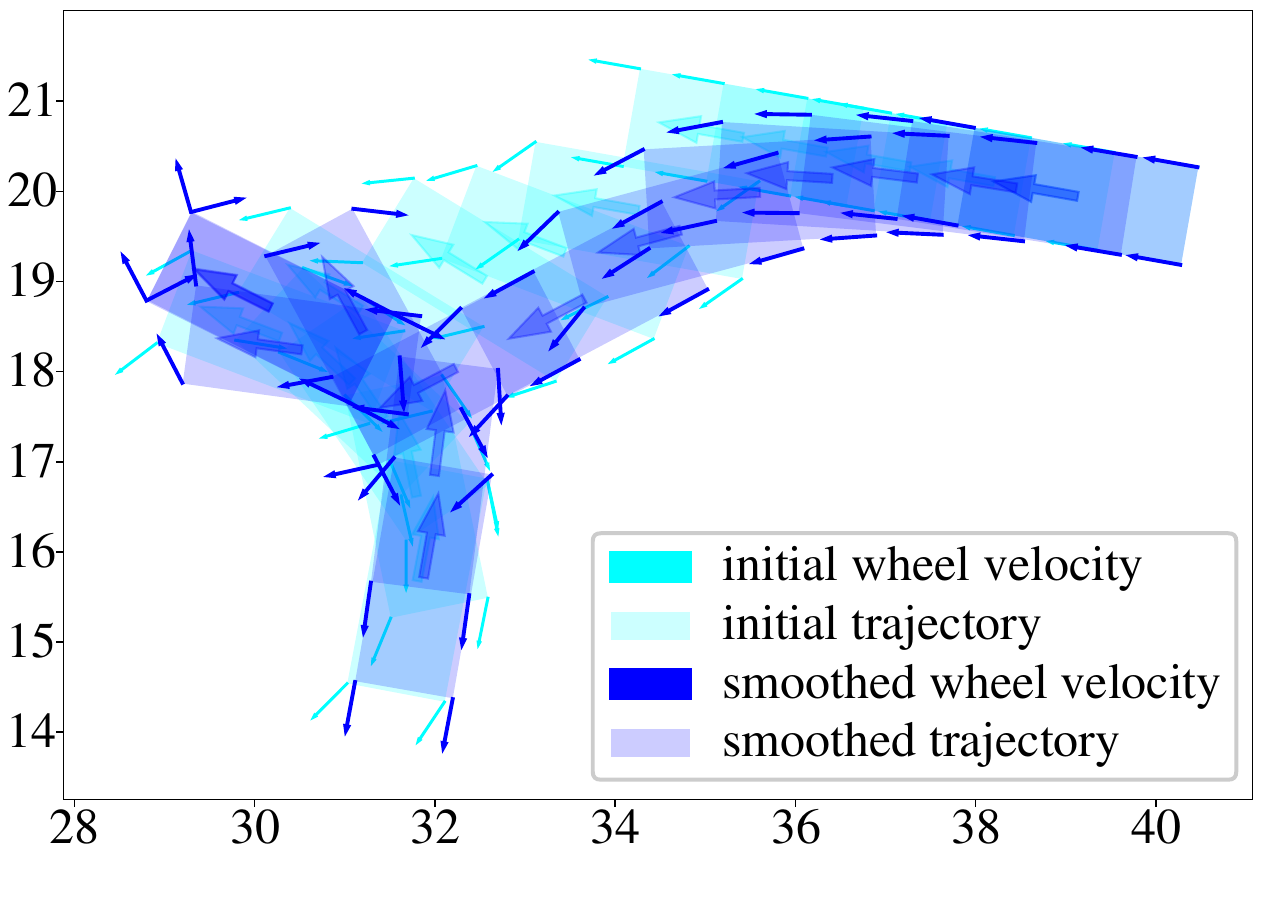}
    \includegraphics[width=0.48\linewidth]{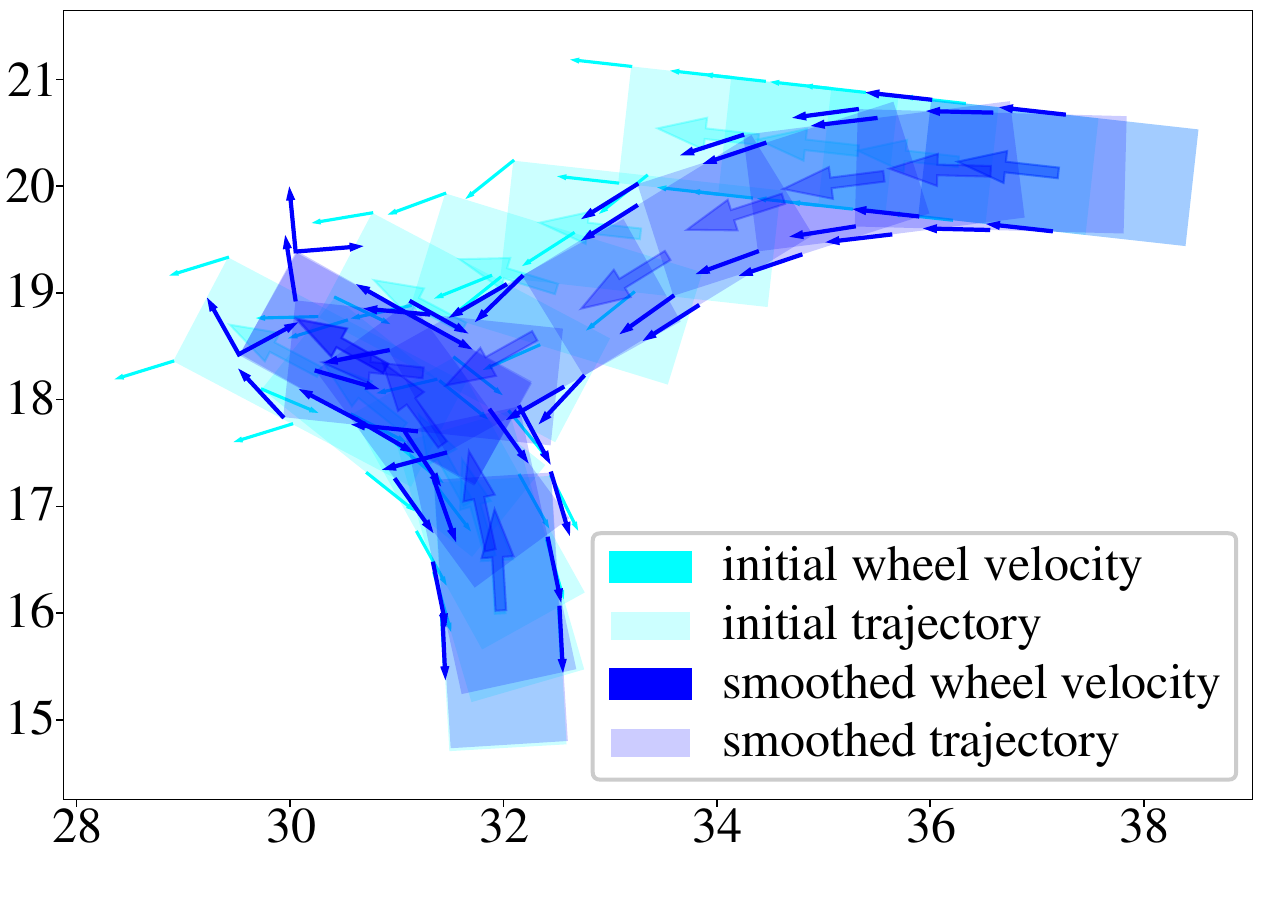} \\
    \vspace{-0.4em}
    \includegraphics[width=0.48\linewidth]{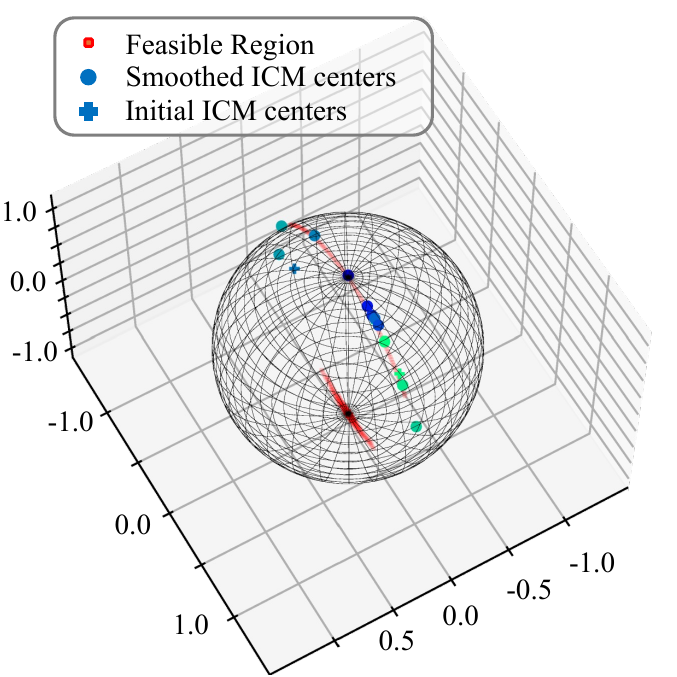}
    \includegraphics[width=0.48\linewidth]{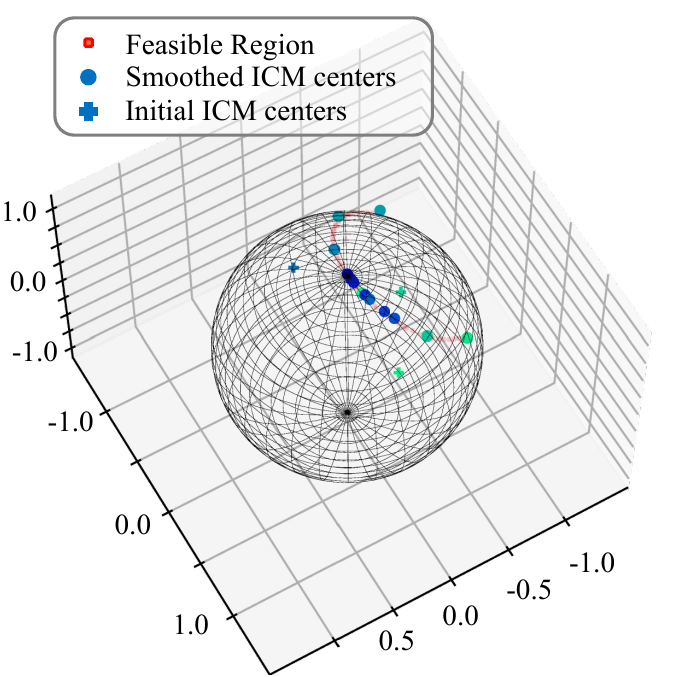}
    \caption{The annotations in this figure is the same as Fig.~\ref{fig:instance_90}. Specially, for \textbf{both} left and right columns, the rear steering limits are set to $\pm0.001$\textdegree~and the front steering limits are both $\pm75$\textdegree. In the \textbf{left} column, the position of the rear wheel to the control center point is symmetrical with the front wheel. In the \textbf{right} column, the position of the rear wheels is moved forward besides the control center.}
    \label{fig:instance_0_half}
    % \vspace{-2.0em}
\end{figure}

To validate the effectiveness of the C-AWS trajectory generator,
% , a series of experimental tests were conducted under diverse conditions and constraints. 
% Fig.~\ref{fig:instance_90}, \ref{fig:instance_0_half} present these results. 
% These experiments included theoretical analysis visualization as well as simulation trials.
we present optimized trajectories with turning angle limitations set at $\pm$90\textdegree~and $\pm$75\textdegree~in Fig.~\ref{fig:instance_90}. 
% The steering angles are set to $\pm$75\textdegree without statement.
Our optimizer can effectively confine the solution within the prescribed turning angle constraints for the wheels, which can be seen from the previous frame approaching the turn. 
The $\pm$90\textdegree~steering limit allows for a larger lateral angle than $\pm$75\textdegree. 
% and fig.~\ref{fig:instance_75}.
The vehicle also successfully changed from a forward movement mode to a backward movement mode, while avoiding the in-situ rotation mode that requires two stops.
Moreover, the optimized trajectories exhibit improved smoothness compared with initial guesses, which can be observed from ICM moving smoother and closer to the vehicle, especially when the vehicle is turning. 

% , particularly when large-angle turns are required for the wheels. 
% This is evident from the reduced number of sharp turns and the smoother transitions between poses. 
% These results demonstrate the robustness and adaptability of our model in handling various constraints and scenarios, thereby confirming its practical utility in real-world applications.

% Path Planning Efficiency: Our hybrid A* path searching and smoothing method was benchmarked against other established planners in a variety of simulated scenarios. The results indicated that our planner achieved competitive or superior pathfinding times while ensuring smoother trajectories and adherence to the wheel steering constraints. The open-sourced nature of the planner allowed for cross-validation by peers and further improvements from the community.

% Physical Implementation: To confirm the practicality of the proposed model and path planner, they were integrated onto a custom-built WMR platform with limited wheel rotation ranges akin to the airport baggage carrier illustrated in Fig.~\ref{fig:wheel_structure}. The physical experiments demonstrated successful navigation and adherence to the specified constraints, proving the applicability of our theoretical framework to real-world scenarios.

We verified that our model is applicable to different forms of chassis layouts with good robustness.
The experiments are conducted by restricting the steering angle range of the rear wheel to $\pm$0.001\textdegree~while limiting the front wheel's angular range to $\pm$75\textdegree~to mimic bicycle models with different wheelbase and control center. 
Experimental results demonstrate that regardless of how we position the wheels 
% at either side~(assuming the control center as the center of the rear axle) or the rear of the vehicle's control center, as shown in fig.~\ref{fig:instance_0_half} 
% and fig.~\ref{fig:instance_0}
, the ICM converges to compliant positions within these constraints as shown in Fig.~\ref{fig:instance_0_half}.
In this experiment, the initial guesses were explored under conditions where both front and rear wheels are constrained to a $\pm$75\textdegree~limit, and the model self-converges to states that adhere to the constraints in subsequent steps. 
This indicates that our model possesses good robustness, as it can tolerate some errors in the initial guess.
In addition, we found that for forward sampling simulation processes, the choice of the control center should ensure that a sufficient number of effective sampling points are available. 
In the case of the bicycle model, setting the control center at the center of the rear axle generally offers a better path to approaching the goal.

\subsection{Trajectory Following Experiments}
\label{sec:following}

\begin{figure}[!t]
    \centering
    \vspace{0.8em}
    % \includegraphics[width=0.9\linewidth]{img/AWSGlobalMassPointOptimizer smoother.pdf}
    % \includegraphics[width=0.7\linewidth]{img/simulator.jpg}\\
    % \vspace{-2.4em}
    \includegraphics[width=0.9\linewidth]{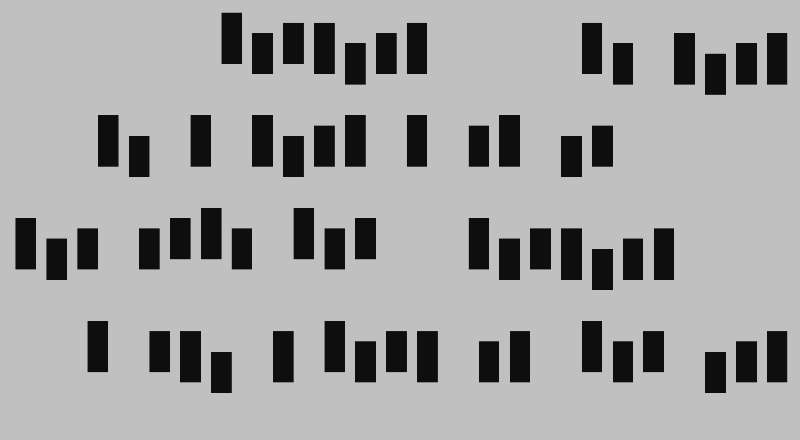}
    \caption{
    % The digital twin of experiment platform~(up) and 
    The map of the experiment parking scene.}
    \label{fig:simulator}
    \vspace{-0.5em}
\end{figure}

We conducted the trajectory following experiments in NVIDIA Isaac Sim~\cite{issac} simulator, which can offer a realistic physics engine and accurate robot status. 
The simulator allows us to thoroughly develop and test our trajectory generator with high repeatability and flexibility.
These evaluations were primarily carried out in a car park environment as Fig.~\ref{fig:simulator} where precise and agile maneuvers are essential.
To minimize the impact of the controller, the path followers are the same as C-AWS trajectory generators except the time horizon is much shorter.
Velocity commands of the vehicle body are subsequently translated into specific velocity and angular commands for each wheel in the Vehicle Control Unit~(VCU) at a rate of around $10~\textup{Hz}$.

\begin{figure}[!t]
    \centering
    \includegraphics[width=\linewidth]{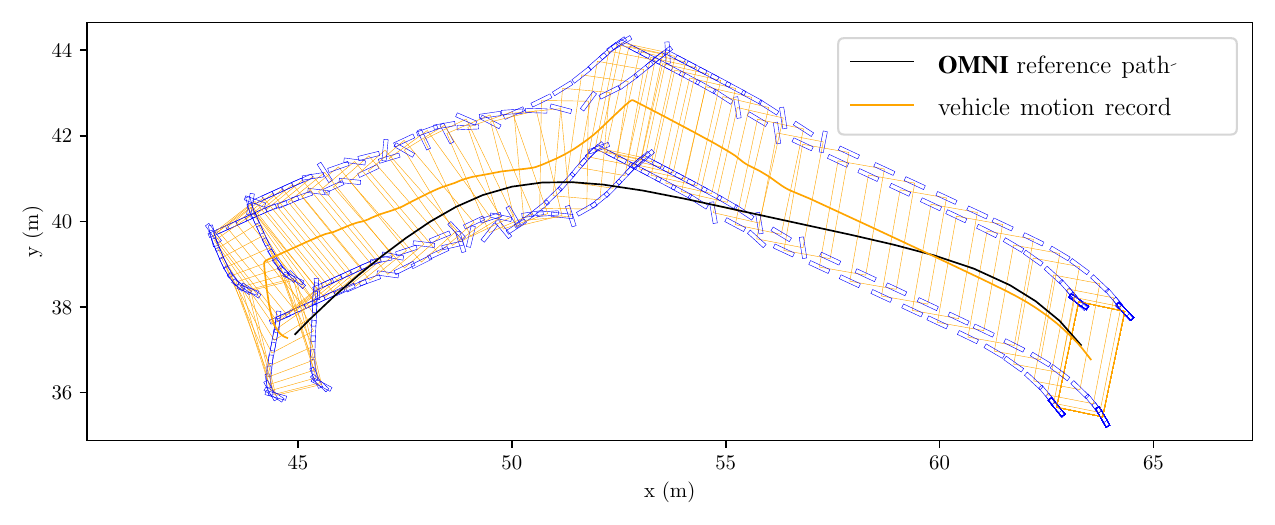}\\
    \vspace{-0.8em}
    \includegraphics[width=\linewidth]{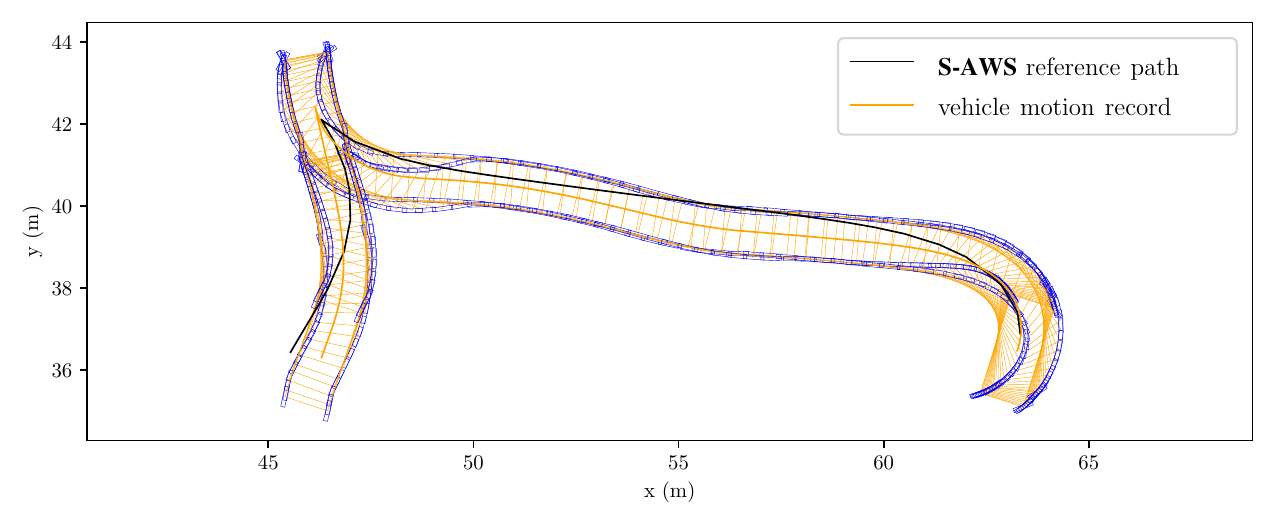}\\
    \vspace{-0.8em}
    \includegraphics[width=\linewidth]{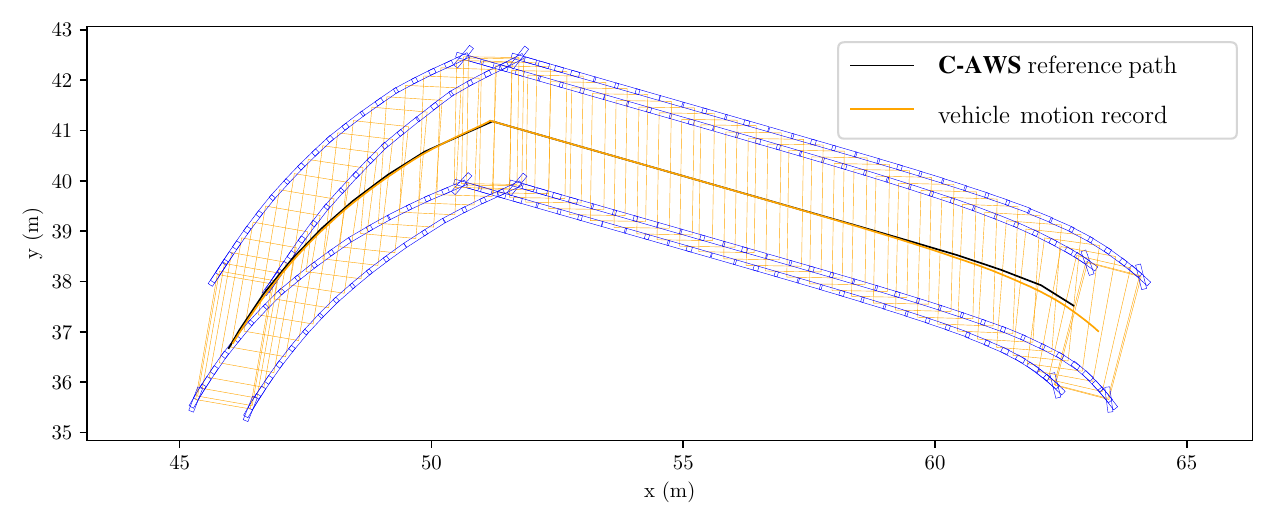}\\
    \vspace{-0.8em}
    \caption{
    Smoothed paths and motion records are depicted in the simulator. 
    The robots are tasked with locomotion from the pose on the right to the left. 
    The black line represents the reference path, while the orange line indicates the recorded control center. 
    Additionally, the orange and blue rectangles illustrate the recorded body and wheel poses, respectively.
    }
    \label{fig:sim_path}
    \vspace{-1.5em}
\end{figure}
\begin{figure}[!t]
    \centering
    \includegraphics[width=\linewidth]{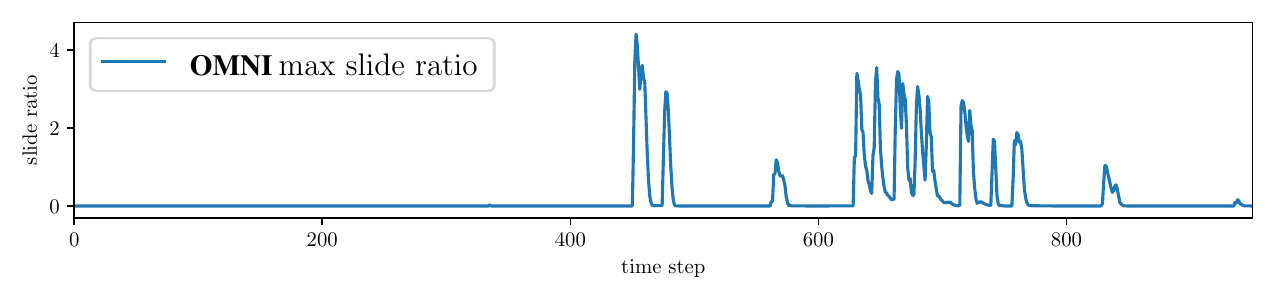}\\
    \vspace{-0.8em}
    \includegraphics[width=\linewidth]{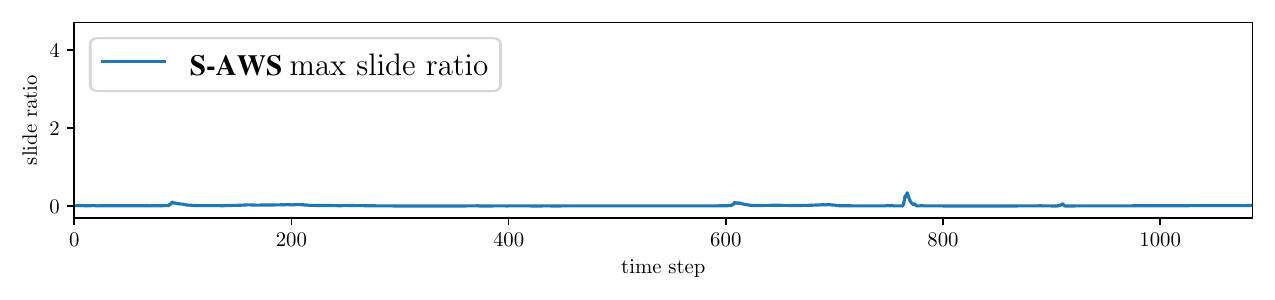}\\
    \vspace{-0.8em}
    \includegraphics[width=\linewidth]{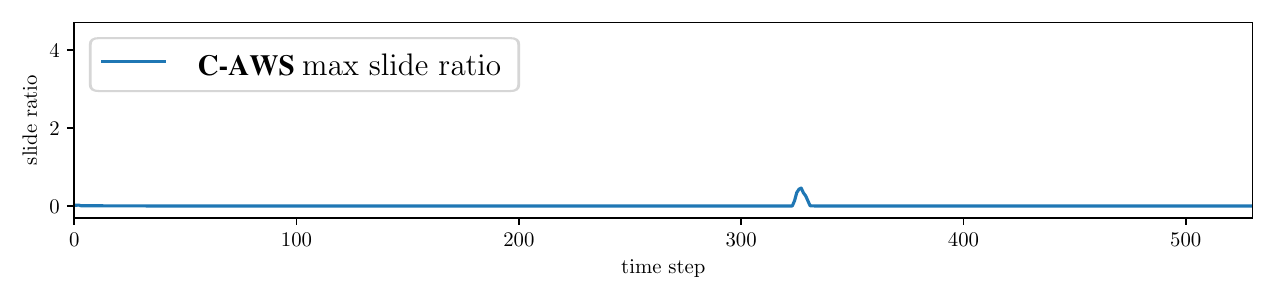}\\
    \vspace{-0.8em}
    \caption{The max slide ratio among wheels during trajectory following process.}
    \label{fig:sim_slide_ratio}
    \vspace{-0.6em}
\end{figure}

In Fig.~\ref{fig:sim_path}, we sequentially compare the motion records of trajectory generators: \textbf{OMNI}directional moving without steering limit constraint, front-rear symmetrical steering AWS~(\textbf{S-AWS}), and our \textbf{C-AWS} generator. 
We have quantitatively assessed the degree of wheel slippage by slide ratio $\lambda$, 
\begin{equation*}
    \lambda = \mathop{\max}_{wheel} \left(v_{wheel}\cdot\cos\left(\delta_{real}^{wheel}-\delta_{ref}^{wheel}\right)\right),
\end{equation*}
where the reference steering $\delta_{ref}$ for each wheel comes from the ideal state the vehicle should be in under the recorded moving speed.
The $\lambda$ along time steps are shown in Fig.~\ref{fig:sim_slide_ratio}.

From the first row in Figs.~\ref{fig:sim_path}~and~\ref{fig:sim_slide_ratio}, it is evident that due to the absence of foreseeing steering constraints in \textbf{OMNI}, the wheels experience substantial unexpected jittering and slippage at discontinuous points in the control space, which in turn negatively impacts both the accuracy of trajectory tracking and the execution speed.
In the second row in Figs.~\ref{fig:sim_path}~and~\ref{fig:sim_slide_ratio}, the optimal trajectory is generated using \textbf{S-AWS}, whose chassis geometry can also be considered as a standard bicycle model with a half-wheel base.
From the figures, there are noticeable trade-offs as the vehicle's posture adjustments consume a significant amount of time, thereby prolonging the overall execution time. 
Indeed, due to the matching rotation angles of both the front and rear wheels of this chassis, it is inherently less susceptible to slipping.
By contrast, our \textbf{C-AWS} optimizer facilitates the robot with more flexibility and completes the route with less time, and lower tracking error, while also exhibiting reduced wheel slippage, which can be readily drawn from the third row in Figs.~\ref{fig:sim_path}~and~\ref{fig:sim_slide_ratio}.

% Furthermore, we have quantitatively assessed the degree of wheel slippage using a formula, which is $\lambda = \max_{wheel} v_{wheel}*|\delta_{real}^{wheel}-\delta_{ref}^{wheel}|$. The reference steering $\delta_{ref}$ for each wheel comes from the ideal state the vehicle should be in under the given motion condition. As depicted in fig.~\ref{fig:sim_slide_ratio}, our method demonstrates a significantly lower occurrence of slippage, allowing for stable control over the vehicle's posture throughout the maneuver.

\subsection{Quantitative Comparison}
%%%%%%%%%%%%%%%%%%%%%%%%%%%%%%%%%%%%%%%%%%%%%
\begin{table*}[t]
\vspace{0.8em}
\centering
\caption{Dynamic Performance Comparison}
\label{tab: dynamic}
\begin{tabular}{ccccccccccccc}
\hline
\multirow{2}{*}{Method} &
  \multirow{2}{*}{$\Delta |d|~\textup{(m)}$} &
  \multirow{2}{*}{$\Delta \theta~\textup{(rad)}$} &
  \multirow{2}{*}{Slide Ratio $\lambda$} &
  \multicolumn{3}{c}{Ave. Velocity~($\textup{m/s}$)} &
  \multicolumn{3}{c}{Ave. Acceleration~($\textup{m/s}^2$)} &
  \multicolumn{3}{c}{Ave. Jerk~($\textup{m/s}^3$)} \\
                 &       &       &                & $v_x$ & $v_y$ & $v_{yaw}$ & $a_x$ & $a_y$ & $a_{yaw}$ & $j_x$ & $j_y$ & $j_{yaw}$ \\ \hline
OMNI             & 0.236 & 0.167 & 0.069$\pm$0.24 & 1.161 & 0.838 & 0.186     & 0.398 & 0.352 & 0.156     & 29.77 & 26.14 & 11.96     \\
S-AWS            & 0.214 & 0.140 & 0.015$\pm$0.08 & 1.463 & 0.068 & 0.187     & 0.305 & 0.157 & 0.145     & 1.261 & 1.200 & 0.640     \\
W/O Smooth       & 0.650 & 0.224 & 0.056$\pm$0.18 & 1.770 & 0.620 & 0.180     & 0.423 & 0.338 & 0.186     & 2.481 & 2.315 & 1.045     \\
C-AWS~(ous)       & 0.154 & 0.108 & 0.008$\pm$0.04 & 1.336 & 0.561 & 0.125     & 0.294 & 0.214 & 0.076     & 1.732 & 1.298 & 0.801     \\
\hline
\end{tabular}
\vspace{-0.8em}
\end{table*}
%%%%%%%%%%%%%%%%%%%%%%%%%%%%%%%%%%%%%%%%%%%%%
\begin{table}[t]
\vspace{0.4em}
\centering
\caption{Operation Time Comparison}
\label{tab: operation time}
\begin{tabular}{ccccc}
\hline
\multirow{2}{*}{Method} &
  \multirow{2}{*}{\begin{tabular}[c]{@{}c@{}}Success \\ Rate\end{tabular}} &
  \multirow{2}{*}{\begin{tabular}[c]{@{}c@{}}Search\\ 20steps~(s)\end{tabular}} &
  \multirow{2}{*}{\begin{tabular}[c]{@{}c@{}}Optimize\\ 20steps~(s)\end{tabular}} &
  \multirow{2}{*}{\begin{tabular}[c]{@{}c@{}}Headless\\ 20steps~(s)\end{tabular}} \\
                 &          &             &             &             \\ \hline
\multirow{2}{*}{OMNI} & \multirow{2}{*}{99.00\%}  & 2.9537      & 0.2467      & 0.1137      \\
                 &          & $\pm$9.7958 & $\pm$0.2083 & $\pm$0.0571 \\
\multirow{2}{*}{S-AWS}    & \multirow{2}{*}{100.00\%} & 0.8038      & 2.1572      & 1.3876      \\
                 &          & $\pm$0.7884 & $\pm$0.4868 & $\pm$0.3268 \\
% w/o smooth       & \text{-} &             & \text{-}    &             \\
%                  &          &             &             &             \\
\multirow{2}{*}{C-AWS~(ours)}             & \multirow{2}{*}{98.00\%}  & 0.5994      & 1.5680      & 0.9678      \\
                 &          & $\pm$1.4477 & $\pm$1.2595 & $\pm$1.0592 \\ \hline
\end{tabular}
\vspace{-1.6em}
\end{table}

We further verify the dynamic performance and computational efficiency of our algorithm. The experiments were conducted in the same setup as Section~\ref{sec:following}. 
Each group of experiments randomly sampled 100 pairs of starting points and ending points in a parking lot environment and then allowed the vehicle to complete tracking of the trajectory under the instructions of the controller. 
% The wheel speed is a speed closed loop and the turning angle is an angle closed loop. 
% The vehicle's speed command issuance uses the same model as smoother, but the horizon is reduced to 3 and the control frequency is about 10Hz.

% 我们对比的方法包括，
% Mass Point Model：为了证明无视车轮转角约束的情况下会对车辆运动造成什么影响，我们屏蔽了搜索和优化过程中有关车轮约束的部分，此时模型退化为质点模型。
% Bicycle Model：将后轮转角约束置为0.001的自行车模型，设置这个实验主要为了证明我们所提出方法的泛化性，能够满足自行车模型这一基本模型。
% w/o smooth：该实验代表我们直接将hybrid A*生成的轨迹给到轨迹跟踪控制器进行跟踪，与该实验的对比主要为了证明我们所提出方法对车辆行驶平滑性和轨迹易跟踪性的contribution。
We quantitatively evaluated the aforementioned trajectory generator and followers, including
\textbf{OMNI:} to prove the impact on vehicle motion if the steering constraint is ignored;
\textbf{S-AWS:} to prove the generalization of our proposed method and that it can satisfy the basic model of the bicycle model;
\textbf{C-AWS:} to prove it really works and illustrate its properties. 
We additionally add baseline \textbf{W/O Smooth} by directly feeding the trajectory generated through Hybrid-A* to the trajectory tracking controller for tracking. The comparison with this baseline is mainly to prove the contribution of our proposed method to the smoothness of vehicle driving and the ease of trajectory tracking. Average tracking error $\Delta|d|$, average heading error $\Delta\theta$, and the maximum slide ratio $\lambda$ among wheels and dynamic performance are listed in Table~\ref{tab: dynamic}.

% 从表1我们可以看出，我们所提出的方法wheel slide rario更低，控制更稳定，行驶过程中对车辆机械损伤较小。
% 并且行驶过程中有更小的加速度，可以更好的保证乘坐人员舒适性和所承载货物的安全性。
% 从表2可以看出，虽然车轮约束会导致较长的轨迹规划时间，但其依然可以基本满足轨迹规划模块的时效需求。

From Table~\ref{tab: dynamic}, we can see that the method we proposed has a lower wheel slide ratio, which can lead to more stable control and less mechanical damage to the vehicle during driving.
And there is smaller acceleration during driving, which can better ensure the comfort of the passengers and the safety of the cargo carried.
By comparing with \textbf{W/O Smooth}, the transformation speed before optimization is faster, but this is achieved by sacrificing the acceleration and deceleration process and traceability performance.
The operation times are listed in Table~\ref{tab: operation time}.
Our method illustrates that computation of wheel constraints will lead to a longer trajectory planning time, they can still basically meet the real-time requirements of the trajectory planning module.
Furthermore, we computed the optimization time excluding the initialization time of optimizer which can be accomplished by programming refraction. 
The results are labeled as \textbf{Headless} and reveals there still retains approximately 40\% operation time reduction potential.

\section{Conclusion}

% In this article, we successfully solved the trajectory planning problem of constrained all-wheel steering robots. 
% By using a search-first-then-optimize framework and reformulating the AWS kinematic model, our approach can overcome the nonlinearity challenges due to steering limits, enhance maneuverability, and improve passenger comfort by minimizing acceleration changes. 
% The method proved superior to non-smooth alternatives and meets real-time operation needs.
% In the future, we will further optimize the running speed of searching and optimization time and combine it with other tasks to evaluate the improvement of this type of chassis for various tasks.
% Besides, C-AWS can be equipped with modern differentiable collision-avoiding algorithms in both static and dynamic scenarios~\cite{han2023differential} ensuring safer and more efficient transporting.

In this work, we addressed the trajectory planning problem for C-AWS robots. We employed a search-first-then-optimize framework, reformulating the AWS kinematic model to tackle the nonlinearity arising from steering limits. 
In this framework, a time-cost-2nd-order initial trajectory planner was introduced. 
It was widely suitable for robots with fixed steering axis positions. 
Additionally, a trajectory optimizer that incorporates our kinematic model was proposed, linearizing steering constraints to generate feasible paths and control mechanisms for C-AWS. 
Our approach not only enhanced maneuverability but also improved passenger comfort by reducing acceleration changes. 
Compared to alternatives, our method was superior and met the requirements for real-time operation.

For future work, we aim to enhance the efficiency of our method by optimizing the running speed of the search and optimization processes. We also plan to integrate our approach with various tasks to assess how this type of chassis can be improved for different applications. Additionally, the C-AWS system will be upgraded with modern differentiable collision-avoidance algorithms~\cite{han2023differential}. These improvements will be applicable in both static and dynamic environments, promising safer and more efficient transportation.

% The experimental section provides empirical evidence supporting the superiority and practical utility of our linearized model and unified cost global path planner for WMRs operating under various constraints, including limited wheel steering angles. The results underscore the potential for wider adoption of such technology in applications where maneuverability, precision, and safety are paramount.

% Compatible with Modern Planning methods, including collision avoidance, and minimum jerk optimization. Sim2Real, simulator contributes to high-risk development. why not reinforcement learning? continuous problem. why no time comparison: no extra computing load, rely on solver
% 对于大雨90的情况来说，车轮的调度需要相对更复杂的调度策略，因此未被讨论

\bibliographystyle{IEEEtran}
\bibliography{IEEEabrv,ref}

% \bibliographystyle{IEEEtran}
% \bibliography{egbib}

\vfill
\end{document}